\documentclass{article}

\usepackage{arxiv}

\usepackage[utf8]{inputenc} 
\usepackage[T1]{fontenc}    
\usepackage{hyperref}       
\usepackage{url}            
\usepackage{booktabs}       
\usepackage{amsfonts}       
\usepackage{nicefrac}       
\usepackage{microtype}      
\usepackage{lipsum}
\usepackage{graphicx}
\usepackage{hyperref}
\usepackage{epstopdf, epsfig}
\usepackage{bm}
\usepackage{amsthm}
\usepackage{amsmath}
\usepackage{amssymb}
\usepackage[utf8]{inputenc}
\usepackage{multirow,booktabs}
\usepackage{tabularx}
\usepackage[T1]{fontenc}
\usepackage{mathptmx}
\usepackage{multirow}
\usepackage[numbers]{natbib}
\usepackage{float}
\usepackage{eucal}
\usepackage{color}
\usepackage{subfigure}
\usepackage[ruled,linesnumbered]{algorithm2e}
\usepackage{natbib}
\usepackage{caption}
\SetKwComment{Comment}{$\triangleright$\ }{}

\title{Multi-fidelity wavelet neural operator}

\author{
  Akshay Thakur \\
  Department of Applied Mechanics\\
  Indian Institute of Technology Delhi\\
  Hauz Khas - 110016, New Delhi, India \\
  \texttt{akshaythakur1482@gmail.com} \\
  \And
  Tapas Tripura\\
  Department of Applied Mechanics\\
  Indian Institute of Technology Delhi\\
  Hauz Khas - 110016, New Delhi, India \\
  \texttt{tapas.t@am.iitd.ac.in}
   \And
 Souvik Chakraborty \\
  Department of Applied Mechanics\\
  School of Artificial Intelligence\\
  Indian Institute of Technology Delhi\\
  Hauz Khas - 110016, New Delhi, India \\
  \texttt{souvik@am.iitd.ac.in} \\
}

\begin{document}
\maketitle

\begin{abstract}
Operator learning frameworks, because of their ability to learn nonlinear maps between two infinite dimensional functional spaces and utilization of neural networks in doing so, have recently emerged as one of the more pertinent areas in the field of applied machine learning. Although these frameworks are quite capable when it comes to modeling complex phenomena, they require an extensive amount of data for successful training which is often not available or is too expensive. However, this issue can be alleviated with the use of multi-fidelity learning, where a model is trained by making use of a large amount of inexpensive low-fidelity data along with a small amount of expensive high-fidelity data. To this end, we develop a new framework based on the wavelet neural operator which is capable of learning from a multi-fidelity dataset. The developed model's learning capabilities are demonstrated by solving different problems which require effective correlation learning between the different fidelities for surrogate construction. The results obtained from this work illustrate the good performance of the proposed framework. 
\end{abstract}

\section{Introduction}\label{S:1}
The advancements in computational methods and scientific machine learning in recent times have not only helped in the accurate modeling of complex physical systems but have also enabled expeditious developments in the application of multi-fidelity (MF) modeling methods to these systems \cite{babaee2020multifidelity,kuya2011multifidelity, zhang2021multi,mahmoudabadbozchelou2021data, fleeter2020multilevel,chakraborty2017surrogate}. In general, a considerable amount of high-fidelity (HF) data is required to train an accurate single-fidelity surrogate model (SM) for complex physical phenomena. However, the cost of obtaining sufficient HF data is usually exorbitant, and frequently, the availability of HF data is limited; for example, generally, limited data is available from some experiments due to their prolonged nature and costly setup. Furthermore, such limited data availability hinders a complete comprehension of the system. On the other hand, in general, low-fidelity data is easily and cheaply available in abundance. However, training an SM using low fidelity (LF) data leads to sizeable inaccuracies.\par

To alleviate these issues, a potential solution lies in the utilization of MF learning techniques which involve the integration of HF and LF data. The main idea behind MF learning is that easily available low-fidelity data, many times, can provide some beneficial knowledge about the tendencies and/or patterns of the system under consideration. This useful knowledge can then be integrated into a model, called an MF model, which then allows the model to provide predictions with improved accuracy, in comparison to a single fidelity model, using only a small amount of HF data \cite{forrester2007multi,meng2020composite}. Furthermore, several different MF learning approaches, which could be employed in a wide variety of situations, exist in the literature. Perhaps some of the noteworthy approaches are response surface models \cite{vitali2002multi}, Gaussian process regression \cite{forrester2007multi, perdikaris2017nonlinear,raissi2017inferring,zhang2020latent,oune2021latent}, polynomial chaos expansions \cite{padron2016multi, yan2019adaptive}, and deep learning \cite{meng2020composite, zhang2021multi,wang2022mosaic,thakur2022deep}. Especially a significant amount of work has been done on the usage of neural networks for multi-fidelity learning recently. This includes utilization of transfer learning-based approaches \cite{chakraborty2021transfer, de2020transfer,de2022neural,song2021transfer}, approaches which concurrently train LF and HF neural networks (NNs) \cite{meng2020composite, zhang2021multi}, and approaches training the different NNs in a successive fashion \cite{liu2019multi}. Furthermore, all of these approaches could be coupled with physics-informed learning \cite{chakraborty2021transfer,meng2020composite,liu2019multi}. Compared to the MF models based on Gaussian processes, NN-based models offer the advantage that they can handle discontinuous functions \cite{meng2020composite}. Furthermore, NNs can approximate complex functions with relative ease. Apart from that, it should also be noted that the LF and the HF data can be obtained from a combination of different sources, for example, coarse mesh and fine mesh numerical simulations, numerical simulations and experiments, or numerical simulations and analytical solutions, to name a few. \par

One of the more immediate developments in the field of scientific machine learning has been the introduction of a new framework, called neural operators, for learning the mapping of operators between two infinite dimensional spaces by pushing global integral operators, in a manner akin to NNs, through activation functions which are nonlinear and local \cite{lu2019deeponet,li2020fourier,li2020neural,tripura2023wavelet,you2022nonlocal, lu2022comprehensive}. A more archetypical usage of neural operators has been in the construction of SMs for PDE solvers. However, training these neural operator frameworks for accurate outputs requires a large amount of high-fidelity data, making the process computationally expensive. Nevertheless, this computational burden could be relieved by using multi-fidelity data for training. More recently, different studies have considered multi-fidelity data for training DeepONets \cite{de2022bi,lu2022multifidelity,howard2022multifidelity}. However, no such effort has been made for other classes of neural operators. The benefit of MF frameworks developed using DeepONets is that they are naturally suitable for irregular domains because they perform pointwise mapping. However, because of this pointwise nature, output from the model corresponds to the value of the intended output field at a single point. This leads to larger training times, and non-trivial data preprocessing. Furthermore, there is a hindrance in scalability to a large number of fidelities as the multiple fidelity data has to be appended to the branch net\cite{lu2022multifidelity}.\par 
Tripura and Chakraborty \cite{tripura2023wavelet} found that the Wavelet Neural Operator (WNO), which they proposed, managed to outperform other neural operator frameworks in the majority of the complex nonlinear operator learning tasks assessed by them. Due to robustness, WNO also finds application in medical elastography \cite{tripura2023elastography} and fault diagnosis \cite{rani2023fault}. Furthermore, WNO is highly effective in learning patterns in images or signals, tackling both complex and smooth geometries, handling families of particularly nonlinear types of PDEs, and learning the frequency with which changes take place in a pattern within a solution domain. Therefore, to increase the learning efficiency and exploit the different available fidelities of data, in the present work, we propose a methodology for enabling multi-fidelity learning with Wavelet Neural Operator (WNO) from small HF and large LF datasets  by using input supplementation and residual-based learning. Besides, decreasing the number of HF samples necessary for training WNO, in comparison to the existing MF models, the proposed methodology offers the following novelties:
\begin{itemize}
    \item First, the propounded framework processes the MF data as an image or a signal, i.e., it performs complete field-to-field mapping instead of performing mapping for each point separately, using convolutions, which helps in scalability to high-dimensional data.
    
    \item Second, the proposed framework has a strong ability to handle discontinuities and spikes because of its wavelet kernel integral layer.
    
    \item Third, a WNO-based bi-fidelity framework is presented for an unsteady nonlinear PDE problem.

    \item Fourth, multiple fidelities can be handled, and the aggregation is scalable to a large number of fidelities as the data is handled as an image and/or signal.

    \item Fifth, the developed model is agnostic to the functional relationship between different fidelities, and it can learn families of linear/nonlinear operators and PDEs.

\end{itemize}

The remainder of the paper is structured as follows: section \ref{S:2} provides an overview of WNO and multi-fidelity learning. Furthermore, it describes the methodologies employed to develop a framework for WNO, which enables it to learn from multi-fidelity data. In section \ref{S:3}, the developed framework is put to the test on several numerical problems, and the results indicating the performance of the proposed framework are presented. Finally, in section \ref{S:4}, the concluding remarks are provided.

\section{Methodology}\label{S:2}
We begin with a brief overview of WNO and multi-fidelity learning. The brief overviews are then followed by a delineation of the multi-fidelity scheme employed for WNO in the current work.

\subsection{Wavelet Neural Operator}
As mentioned in section \ref{S:1}, neural operators can learn the operator mappings between two infinite-dimensional function spaces. In general, for a given PDE, the mappings are learned from functions in the input function space such as source term, initial condition, or boundary condition to the PDE solution function in the output function space.\par

For proper comprehension, let us consider an $n$-dimensional fixed domain $D \in \mathbb{R}^{n}$ with boundary $\partial D$. Let, $\boldsymbol{a}: D \mapsto a(x \in D) \in \mathbb{R}^{d_{a}}$ and $\boldsymbol{u}: D \mapsto u(x \in D) \in \mathbb{R}^{d_{u}}$ be the input and output functions in corresponding Banach spaces $\mathcal{A}:=C\left(D ; \mathbb{R}^{d_{w}}\right)$ and $\mathcal{U}:=C\left(D ; \mathbb{R}^{d_{u}}\right)$. Then the nonlinear differential operator to be learned is given by,

\begin{equation}
    \mathcal{D}: \mathcal{A} \ni \boldsymbol{a}(x) \mapsto  \boldsymbol{u}(x) \in \mathcal{U}.
    \label{equation1}
\end{equation}
Considering we have datasets available in the form of $m-$point discretized $N$ pairs of input and output functions $\left\{a_{j} \in \mathbb{R}^{m \times d_{a}}, u_{j} \in \mathbb{R}^{m \times d_{u}}\right\}_{j=1}^{N}$, the operator could be approximated using a NN as follows:

\begin{equation}
    \mathcal{D}: \mathcal{A} \times \boldsymbol{\theta} \mapsto \mathcal{U},
    \label{equation2}
\end{equation}
where $\boldsymbol{\theta}$ denotes NN's finite-dimensional parameter space. The desired neural network architecture for operator learning is achieved by firstly lifting the inputs $a(x) \in \mathcal{A}$ to some high dimensional space by the usage of a local transformation given by $L(a(x)): \mathbb{R}^{d_{a}} \mapsto \mathbb{R}^{d_{v}}$. The transformation in the current case is achieved with the help of a shallow fully connected NN (FNN). In addition, the high dimensional space is denoted as $v_{0}(x)=L(a(x))$ and it only takes values in $\mathbb{R}^{d_{v}}$. Furthermore, the number of total updates or steps required for attaining acceptable convergence is denoted by $l$. Thus, the updates applied on $v_{0}(x)$, can be represented as, $v_{j+1}=F\left(v_{j}\right) \forall j=1, \ldots, l$  where the function $F: \mathbb{R}^{d_{v}} \mapsto \mathbb{R}^{d_{v}}$ only takes value in $\mathbb{R}^{d_{v}}$. After the updates, another local transformation denoted by $L\left(v_{l}(x)\right): \mathbb{R}^{d_{v}} \mapsto \mathbb{R}^{d_{u}}$ is employed to transform the output  $v_{l}(x)$ in the high dimensional to the solution space $u(x)=M\left(v_{l}(x)\right)$. Also, the definition of the step-wise updates $v_{j+1}$ can be provided as follows:

\begin{equation}
    v_{j+1}(x):=f\left(\left(K(a ; \phi) * v_{j}\right)(x)+W v_{j}(x)\right) ; \quad x \in D, \quad j \in[1, l]
    \label{equation3}
\end{equation}
where the non-linear activation function is denoted by $f(\cdot): \mathbb{R} \rightarrow \mathbb{R}$, the linear transformation is represented using $W: \mathbb{R}^{d_{v}} \rightarrow \mathbb{R}^{d_{v}}$, and the integral operator on $C\left(D ; \mathbb{R}^{d_{v}}\right)$ is denoted as $K: \mathcal{A} \times \boldsymbol{\theta} \rightarrow \mathcal{L}(\mathcal{U}, \mathcal{U})$. Furthermore, as our framework is based on neural networks, so $K(a ; \phi)$ is represented as a kernel integral operator parameterized by $\phi \in {\bm {\theta}}$. Thus, $*$ here is the convolution operator. Additionally, with the employment of the degenerate kernel concept, Eq. \eqref{equation3} allows us to learn the mapping between any two infinite dimensional spaces. Finally, for obtaining the WNO framework, the learning of the kernel is performed by parameterizing the kernel in the wavelet domain. Furthermore, a notably essential component of WNO is the wavelet kernel integral layer, a pictorial description of which can be found in Fig. \ref{fig:1}.

\subsection{Multi-fidelity modeling}
The major theme in multi-fidelity learning is the exploitation of correlation between the HF data, which is quite accurate but available in a smaller amount, and the LF data, which is inaccurate but is available in a larger amount. A popular autoregressive strategy for multi-fidelity learning \cite{kennedy2000} can be expressed as follows:

\begin{equation}
    y_{H}=\rho(x) y_{L}+\delta(x),
    \label{equation4}
\end{equation}
where the LF and HF data are respectively denoted by $y_{L}$ and $y_{H}$, $\rho(x)$ is a multiplicative factor that determines the correlation between the two fidelities, and $\delta(x)$ quantifies the corresponding additive correlation. However, the issue with this strategy is that it only captures the linear correlation between the LF and the HF data. In order to account for the nonlinear correlation, Meng and Karniadakis \cite{meng2020composite} proposed a generic autoregressive scheme, which is as follows:

\begin{equation}
    y_{H}=G\left(y_{L}\right)+\delta(x),
    \label{equation5}
\end{equation}
where $G(\cdot)$ is a function, linear or nonlinear and not known a priori, that defines the mapping between the two fidelities. Further, Eq. \eqref{equation5} can also be expressed as,

\begin{equation}
    y_{H}=\mathcal{G}\left(x, y_{L}\right) .
    \label{equation6}   
\end{equation}

\subsection{Proposed Framework}
In the current work, we aim to approximate operator $\mathcal{G}$ in Eq. \eqref{equation6} using the WNO framework. To accomplish this, a small HF dataset, $\mathcal{Q}_{H}$, containing $N_{H}$ number of HF input and output function pairs is generated along with a large LF dataset, $\mathcal{Q}_{L}$,  containing $N_{L}$ pairs of LF input and output function. Once more, the reason for small and large sizes is computational cost. Mathematically, the datasets can be represented as follows:

\begin{align}
    \mathcal{Q}_{H} = \left\{a_{j}^{H}, u_{j}^{H} = \mathcal{D}_{H}(a_{j}^{H}) \right\}_{j=1}^{N_{H}};  \quad \mathcal{Q}_{L} = \left\{a_{j}^{L}, u_{j}^{L} = \mathcal{D}_{L}(a_{j}^{L}) \right\}_{j=1}^{N_{L}},
\end{align}
where the HF operator we want to learn is denoted by $\mathcal{D}_{H}$, the LF operator is denoted using $\mathcal{D}_{L}$, and $N_{H} \ll N_{L}$.\par

Furthermore, to enable WNO for multi-fidelity learning, we follow a two-step approach as shown in Fig. \ref{fig:1}. Firstly, we train a WNO network (LF-WNO) using LF data. Of course, because of the vast supply of LF data, training a low-fidelity surrogate model (LFSM) to a high degree of accuracy is a straightforward task. However, in the second step where we train a separate WNO network (HF-WNO) to learn the HF solution $\mathcal{D}_{H}(a)$, additional strategies such as supplementation of WNO inputs with LF outputs $\mathcal{D}_{L}(a)(x)$ and residual operator learning have to be used.  These approaches are described in the sections below.

\subsubsection{Supplementing HF-WNO inputs with low-fidelity solution}
As established in the previous section, the main goal of multi-fidelity learning is uncovering and exploiting the correlation between LF solution $\mathcal{D}_{L}(a)(x)$ and HF solution $\mathcal{D}_{H}(a)(x)$. To enable the network to effectively learn the unknown operator $\mathcal{G}$ which maps $x$ and $\mathcal{D}_{L}(a)(x)$ to $\mathcal{D}_{H}(a)(x)$, in a data-driven fashion, we augment the inputs  to the HF-WNO network with $\mathcal{D}_{L}(a)(x)$.

\subsubsection{Learning the residual operator}
Instead of training HF-WNO to directly learn the HF operator $\mathcal{D}_{H}$, in residual-based learning, the focus is on learning the residual operator. Residual is nothing but the difference between HF and LF solution, and mathematically, this could be expressed as,

\begin{equation}
    \mathcal{R}^{\dagger}(a)(x) = \mathcal{D}_{H}(a)(x) - \mathcal{D}_{L}(a)(x),
    \label{equation7}   
\end{equation}
where $\mathcal{R}^{\dagger}$ is the residual operator. In general, a correlation might exist between HF and LF solutions. Therefore, a feature similitude, although not exact, could be expected between them. However, note that a correlation might be low and a bias may exist \cite{foumani2022multi}. Simply, the LF solution, despite being inaccurate, preserves the rudimentary feature structure of the HF solution. Therefore, it is a comparatively easier task to learn the residual rather than the HF solution itself.

\subsubsection{Complete Framework}
\begin{figure}[ht!]
    \centering
    \includegraphics[width=1\textwidth]{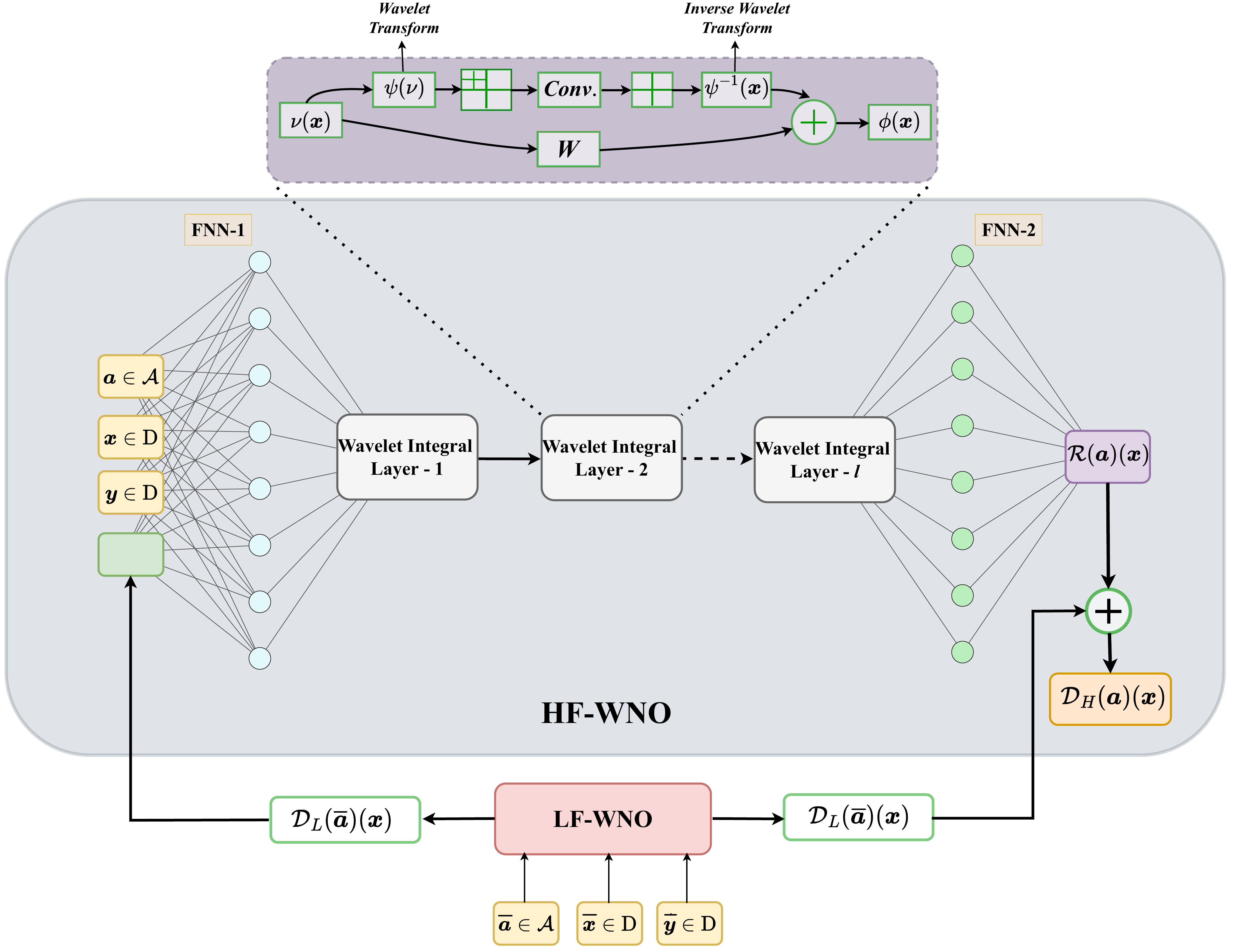}
    \caption{The proposed framework for multi-fidelity wavelet neural operator (MF-WNO). The LF-WNO block is trained beforehand using cheaply available large LF-dataset $\mathcal{Q}_{L}$. The HF-WNO block is subsequently trained using the small-sized HF-dataset $\mathcal{Q}_{H}$ and supplementing the HF-input ${a,x,y}$ with the LF-output $\mathcal{D}_{L}(\bar{a})(x)$ obtained from querying the LF-WNO at LF-inputs ${\bar{a},\bar{x},\bar{y}}$ corresponding to HF-input field $a$. Also, $\mathcal{D}_{L}(\bar{a})(x)$ is subtracted from the high fidelity target $\mathcal{D}_{H}(a)(x)$ to reformulate the training process into learning of residual operator $\mathcal{R}(a)(x)$.}
    \label{fig:1}
\end{figure}

To finally arrive at the complete framework for multi-fidelity WNO, we combine the two-step SM approach with input supplementation and residual operator learning. The pictorial representation of the complete framework is provided in Fig. \ref{fig:1}. In practice, we can directly replace the LF-WNO block with an LF-solver block when we have access to an \textit{efficient LF-solver}.  However, in the absence of an efficient LF-solver, the LF-WNO is first trained using the dataset $\mathcal{Q}_{L}$. As the size of the low-fidelity dataset, $\mathcal{Q}_{L}$ ($N_{L}$) is substantial, it is possible to train LF-WNO with a high level of accuracy.\par 
In the second step, we train the HF-WNO by making use of samples from  $\mathcal{Q}_{H}$ and concatenating the output $\mathcal{D}_{L}(a)(x)$ obtained from querying the LF-WNO or LF-Solver (contingent on the efficiency of LF-WNO) at LF-input field $\bar{a}_j^L$ corresponding to the HF-input field $a_j^H$. Also, the input field  $\bar{a}_j^L$ could be a coarsened instance of $a_j^H$ or it could be the same as $a_j^H$ if the MF learning is performed between different parameters. As shown in Figure \ref{fig:1}, an additional input, the LF solution $\mathcal{D}_{L}(\bar{a})(x)$, is supplied to the network along with $\left\{ a(x,y), x\in D, y \in D\right\}$. It is crucial to note that LF-WNO or the LF-Solver are not trained along the HF-WNO block in the MF-WNO model, i.e., they are not subject to back-propagation, and it is a sequential two-step process. Using a shallow FNN, these inputs are lifted to a higher dimension. The high dimensional output from FNN is then passed through several wavelet kernel integral layers. In the kernel integral layer, to obtain the wavelet coefficients, the inputs first undergo multilevel wavelet decomposition $\psi$. The wavelet coefficients in the sub-band of the last level are then convoluted with the neural network weights, and this is followed by an inverse wavelet transform $\psi(\cdot)^{-1}$ on the convoluted output from the previous step in order to bring the dimensions back to lifted inputs. Simultaneously, in parallel to the kernel integral layer, a local linear transform $W$ is applied to the lifted inputs. The output from the local linear transform is then added to that from the kernel integral layer, and a suitable activation function is applied. After repetition of the same process through the rest of the wavelet kernel integral layers, the output is passed through another FNN, which transforms them back to the intended target output, which is the residual $\mathcal{R}(a)(x)$. Finally, the LF solution is added back to the residual to obtain the desired high-fidelity solution $\mathcal{D}_{H}(a)(x)$.\par

Furthermore, given a suitable loss function, $\mathcal{L}(\mathcal{U}, \mathcal{U})$ is selected, the training of the network can be represented as the following minimization problem:

\begin{equation}
    \boldsymbol{\theta}=\underset{\boldsymbol{\theta}}{\arg \min }\;\mathcal{L}(\mathcal{R}^{\dagger}(\boldsymbol{a}), \mathcal{R}(\boldsymbol{a}, \boldsymbol{\theta})) 
    \label{equation8} 
\end{equation}

\section{Numerical Implementation and Results}\label{S:3}
In this section, we analyze the performance of the proposed approach on eight different example problems. The first five are artificial benchmark problems, the sixth problem is a stochastic ordinary differential equation, the seventh problem is a stochastic partial differential equation (SPDE) on an irregular domain, and the final problem is also an SPDE, but we also demonstrate the uncertainty propagation capabilities of the proposed framework on this problem. Finally, we also demonstrate the usage of the proposed technique for unsteady cases. In addition, for evaluating the accuracy of the proposed approach, we use three metrics for error quantification, specifically, mean-squared error, absolute error, and  coefficient of determination ($R^{2}$-score). Firstly, the evaluation of mean-squared error is done as,

\begin{equation}
    MSE_{\,\text{MFSM}} = L\left({\mathbf{y}}_{MFSM}, {\mathbf{y}}_{HF}\right) = \frac{1}{N}\sum_{i=1}^{N}\left(\mathbf{y}_{{HF}}^{(i)}-\mathbf{y}_{MFSM}^{(i)}\right)^{2},
    \label{equation9}
\end{equation}
where $\mathbf{y}_{MFSM}$ is the solution predicted by the multi-fidelity surrogate model (MFSM) and $y_{HF}$ is the high-fidelity solution. Secondly, the $R^{2}$-score is computed as follows:

\begin{equation}
    R^{2}=1-\frac{\sum_{i=1}^{N}\left(\mathbf{y}_{{HF}}^{(i)}-\mathbf{y}_{MFSM}^{(i)}\right)^{2}}{\sum_{i=1}^{N}\left(\mathbf{y}_{HF}^{(i)}-\bar{\mathbf{y}}_{HF}\right)^{2}}, \quad \text{with } \bar{\mathbf{y}}_{{HF}}=\frac{1}{N} \sum_{i=1}^{N} \mathbf{y}_{HF}^{(i)}.
    \label{equation10}
\end{equation}
Finally, we also perform comparisons of the proposed framework with MF DeepONet for each example problem.

\subsection{Problem Set I: Artificial Benchmarks}
\subsubsection{Artificial Benchmark I: 1-dimensional problem, correlation with input}
\begin{figure}[htbp!]
    \centering
    \includegraphics[width=0.6\textwidth]{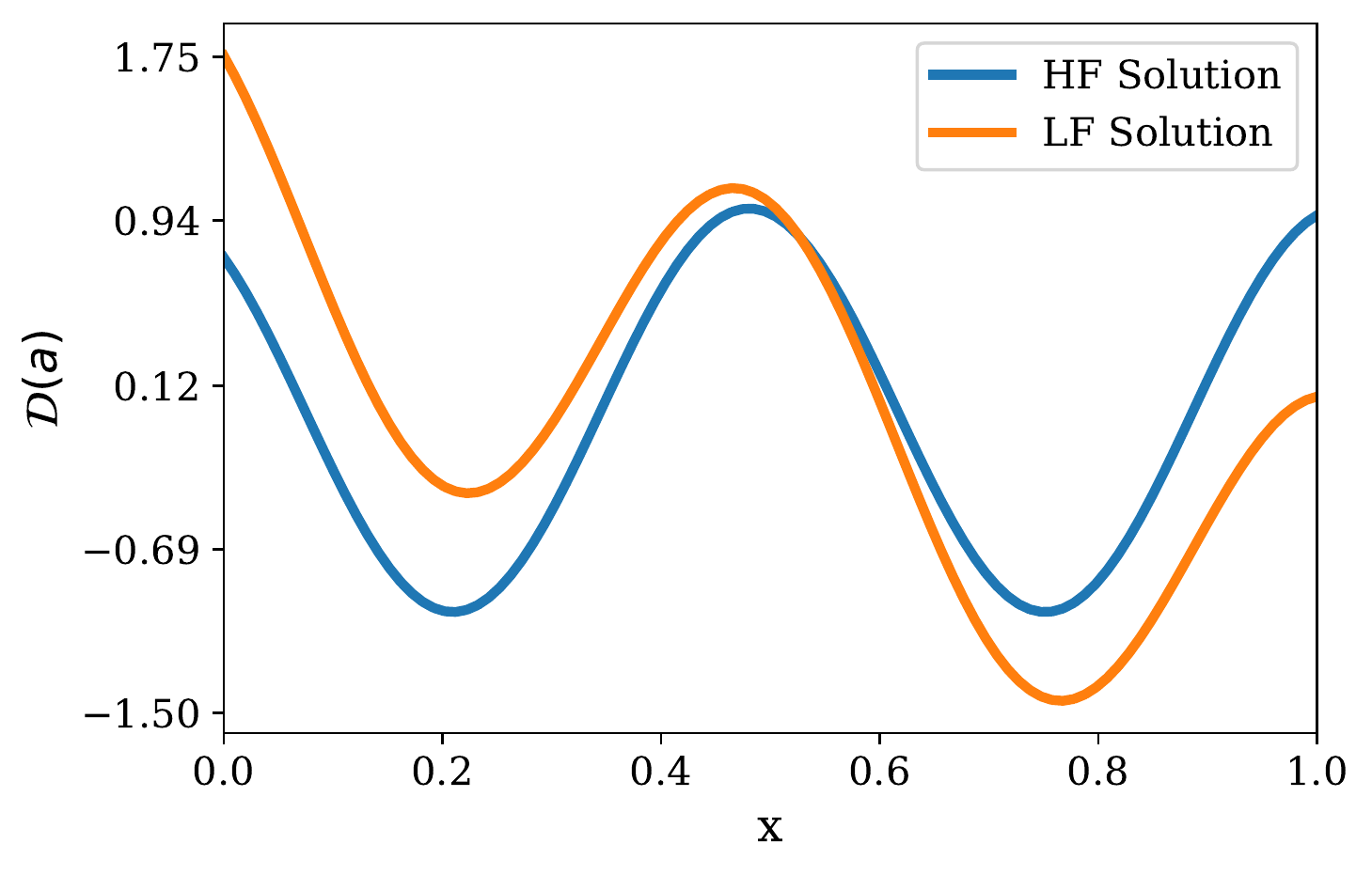}
    \caption{Plots for the exact HF solution and LF solution for 1-d problem having a correlation with $a$.}
    \label{fig:my_label}
\end{figure}

\begin{figure}[htbp!]
    \centering
    \subfigure[]
    {\label{subfig:lab21}\includegraphics[width=0.7\textwidth]{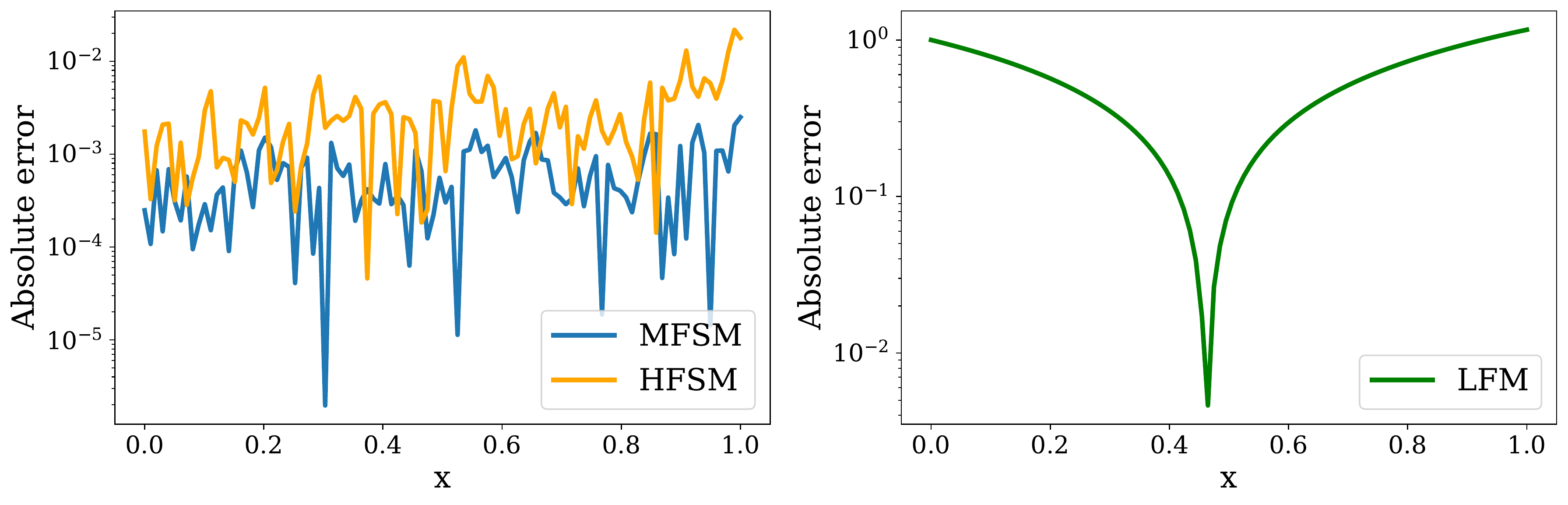}}
    \subfigure[]
    {\label{subfig:lab22}\includegraphics[width=0.7\textwidth]{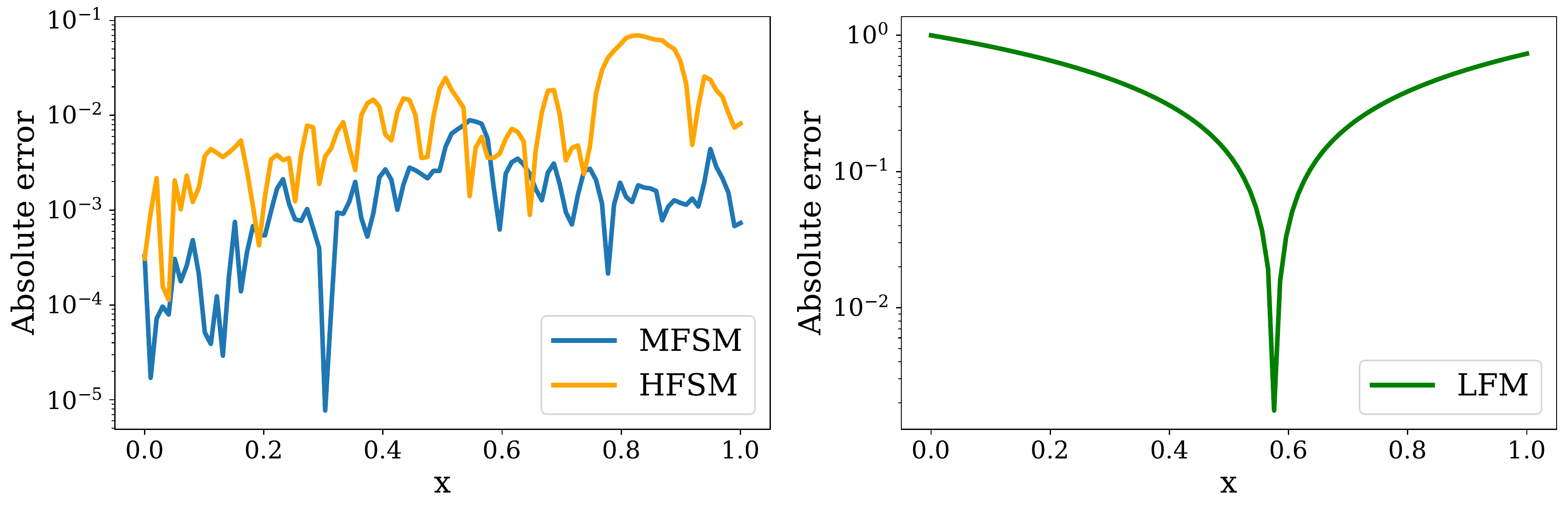}} 
    \subfigure[]
    {\label{subfig:lab23}\includegraphics[width=0.7\textwidth]{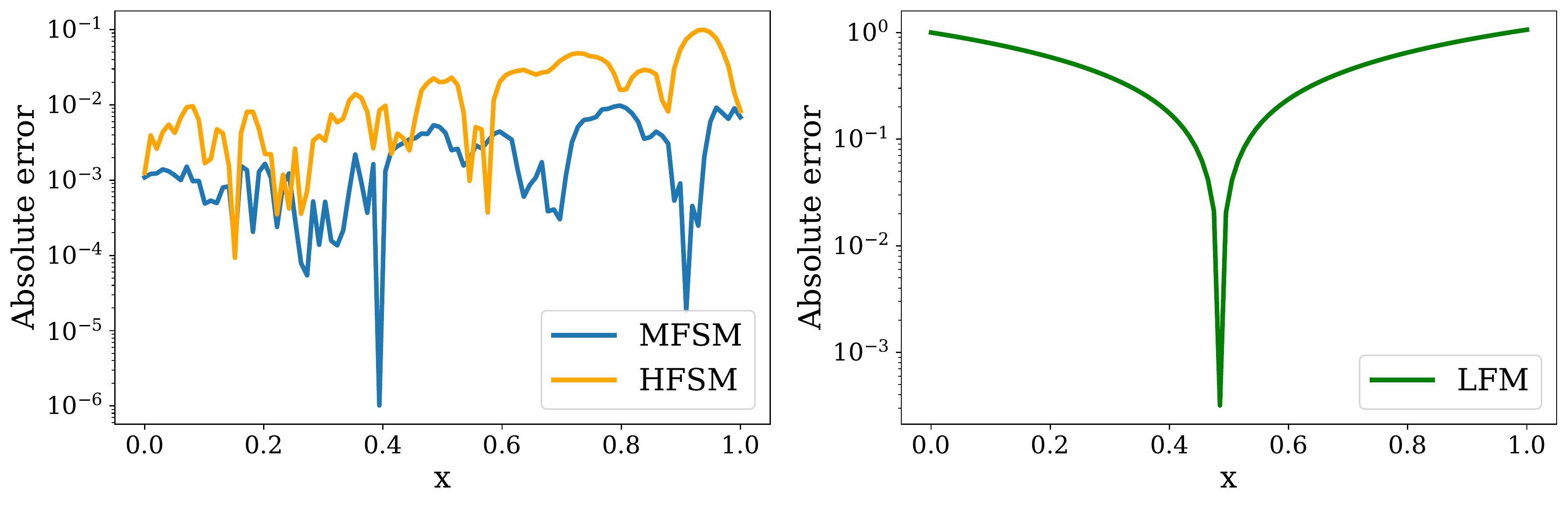}}
    \subfigure[]
    {\label{subfig:lab24}\includegraphics[width=0.7\textwidth]{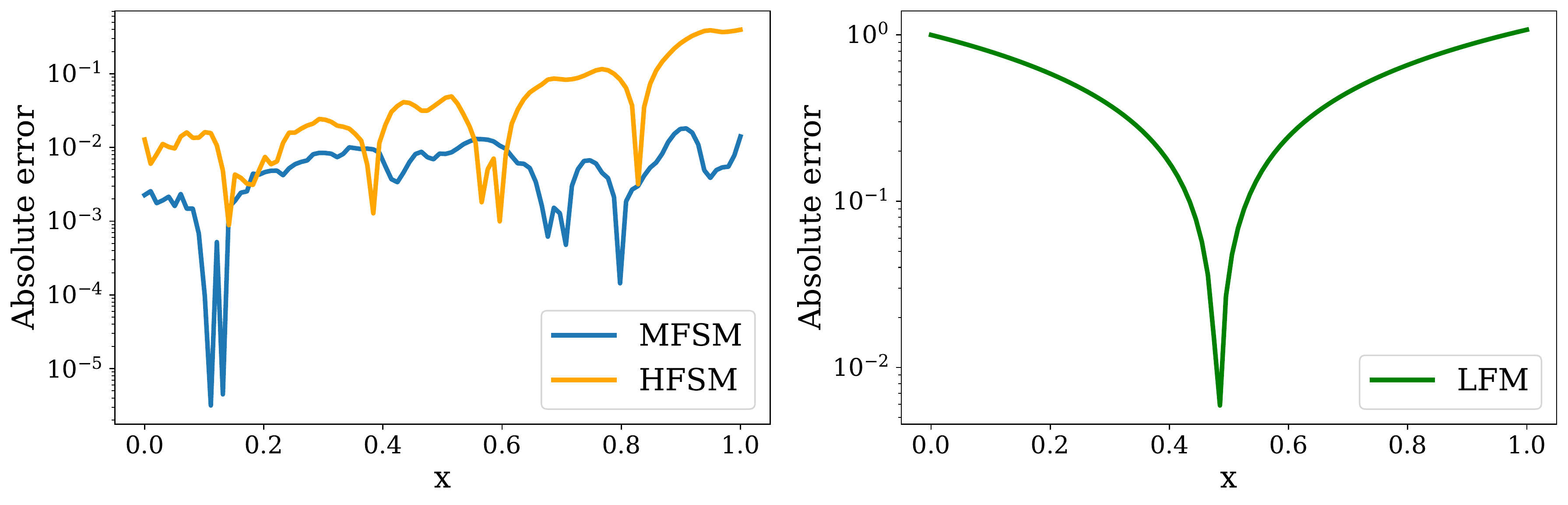}}
    \caption{Absolute error between exact HF solution and prediction from MFSM-WNO and HFSM-WNO on an unseen test set along with absolute error between exact HF and LF solution for training dataset sizes of \textbf{(a)} $10$, \textbf{(b)} $6$, \textbf{(c)} $4$, and \textbf{(d)} $2$ training samples for 1-d problem having a correlation with $a$. The first column shows the absolute error between the exact HF solution and predictions from MFSM and HFSM, and the second column shows the absolute error between LF and the exact solution HF solution. The y-axis has a logarithmic scale.}
    \label{fig:2}
\end{figure}

To demonstrate the multi-fidelity learning abilities of the propounded framework, we consider a problem where the correlation between the HF and LF solution is contingent upon the input function, $a(\cdot)$:

\begin{equation}
    \begin{aligned}
        \mathcal{D}_{L}(a)(x) &=\sin (a)+x-0.25 a, \\
        \mathcal{D}_{H}(a)(x) &=\sin (a), \\
        a &=k x-4,
    \end{aligned}
\end{equation}
where $k \in[10,14]$ and  $x \in[0,1]$. Furthermore, we can represent the equation for the HF solution as follows:

\begin{equation}
    \begin{aligned}
    \mathcal{D}_{H}(a)(x) &=\mathcal{D}_{L}(a)(x)-x+0.25 a\\
                          &=\mathcal{D}_{L}(a)(x)-x+0.25(k x-4). 
    \end{aligned}
\end{equation}
Also, we have presented the actual LF and HF solutions in Fig. \ref{fig:my_label}. Furthermore, an MFSM is trained using HF-WNO (we refer to the trained surrogate as MFSM-WNO) on a multi-fidelity dataset (where LF output solution is concatenated with the HF inputs) of different sizes, which, in this case, are $2$, $4$, $6$, and $10$. Also, for comparison purposes, we train a high-fidelity surrogate model (HFSM-WNO), which is a single-fidelity model, using vanilla WNO on single-fidelity HF datasets of similar sizes, i.e., the number of HF samples used in training HFSM is same as MFSM and no input augmentation or residual learning using LF-solution is present. Furthermore, we also train an MFSM using MF-DeepONet for comparison (we refer to it as MFSM-DeepONet). In addition, we provide MSE on test datasets for models trained using different training dataset sizes along with the MSE between HF and LF solution, $L\left({\mathcal{D}}_{L}(a)(x), {\mathcal{D}}_{H}(a)(x)\right)$ or $MSE_{\text{LF}}$ in Table \ref{tab:1}. Apart from that, we also present the absolute error plots between the exact HF solution and the predicted solution from HFSM and MFSM, along with absolute errors between the exact HF solution and LF solution in Fig. \ref{fig:2}.
\begin{table}[htbp!]
    \centering
    \caption{MSE error between exact HF solution and predictions from  (MFSM-WNO), HFSM-WNO, and MFSM-DeepONet on unseen test dataset for different training dataset sizes for 1-dimensional problem having a correlation with input.}
    \label{tab:1}
    \begin{tabular}{m{3cm}m{2cm}m{2cm} c} 
    \toprule
    \multirow{2}{*}{$\mathcal{Q}_{train}$ Size} &\multicolumn{3}{c}{MSE}\\ \cline{2-4}
    & MFSM-WNO & HFSM-WNO & MFSM-DeepONet \\ [0.5ex] 
    \midrule
    \vspace{0.2em}
    $10$ & 8.8132 $\times 10^{-7}$ & 2.2164 $\times 10^{-5}$ &$4.3338\times 10^{-6}$\\ 
    $6$ & 8.9857 $\times 10^{-6}$ & 3.5422 $\times 10^{-4}$  &$2.5583 \times 10^{-4}$\\ 
    $4$ & 5.9807 $\times 10^{-5}$ & 1.9693 $\times 10^{-3}$  &$1.4725 \times 10^{-3}$\\ 
    $2$ & 4.8607 $\times 10^{-4}$ & 3.4449 $\times 10^{-2}$  & $4.7949 \times 10^{-3}$\\ 
    \hline
    $MSE_{LF}$ & \multicolumn{3}{c}{3.4780 $\times 10^{-1}$} \\
    \bottomrule
    \end{tabular}
\end{table}
Clearly, as evident from Table \ref{tab:1}, the MFSM-WNO, while predicting on the test dataset, is able to outperform the MFSM-DeepONet and HFSM-WNO. The latter by two orders of magnitude. Also, MFSM-WNO  provides a significant improvement over the LF output. Furthermore, Fig. \ref{fig:2} shows the consistently outstanding performance of MFSM-WNO against HFSM-WNO on a single test case for different training dataset sizes. Therefore, it can be stated the MFSM-WNO does a good job of capturing the desired correlation. Finally, we assess the computational cost associated with the developed for inferring the solution of $200$ samples. In this example, the MFSM-WNO requires $0.012$ seconds for predicting the solution for $200$ samples.  

\subsubsection{Artificial Benchmark II: 2-dimensional problem with a nonlinear correlation}
\begin{figure}[htbp!]
    \centering
    \subfigure[]
    {\label{subfig:lab41}\includegraphics[width=0.9\textwidth]{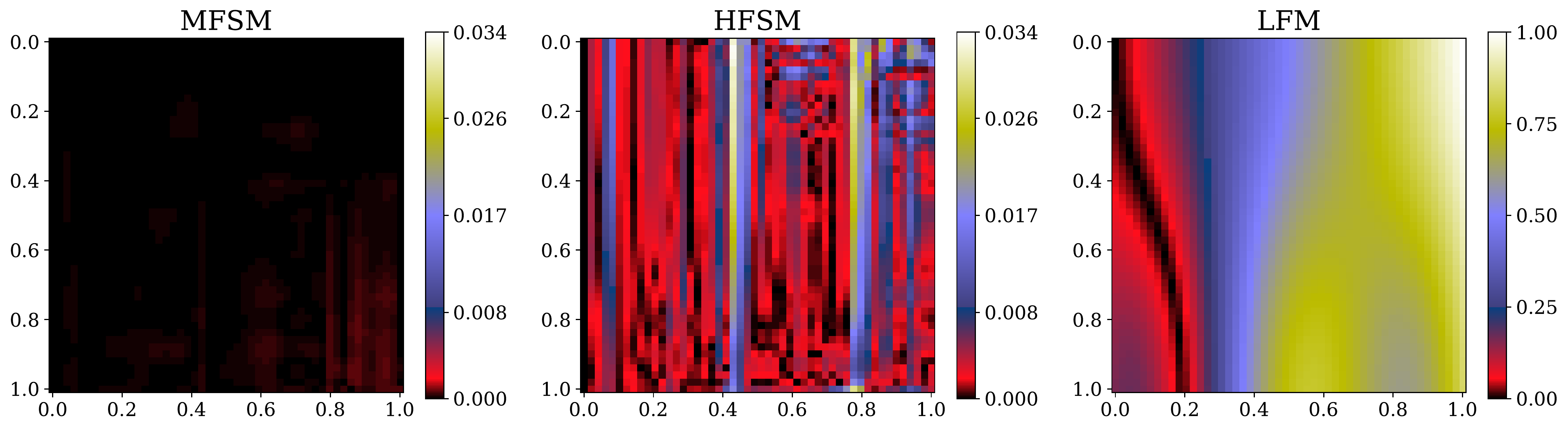}}
    \subfigure[]
    {\label{subfig:lab42}\includegraphics[width=0.9\textwidth]{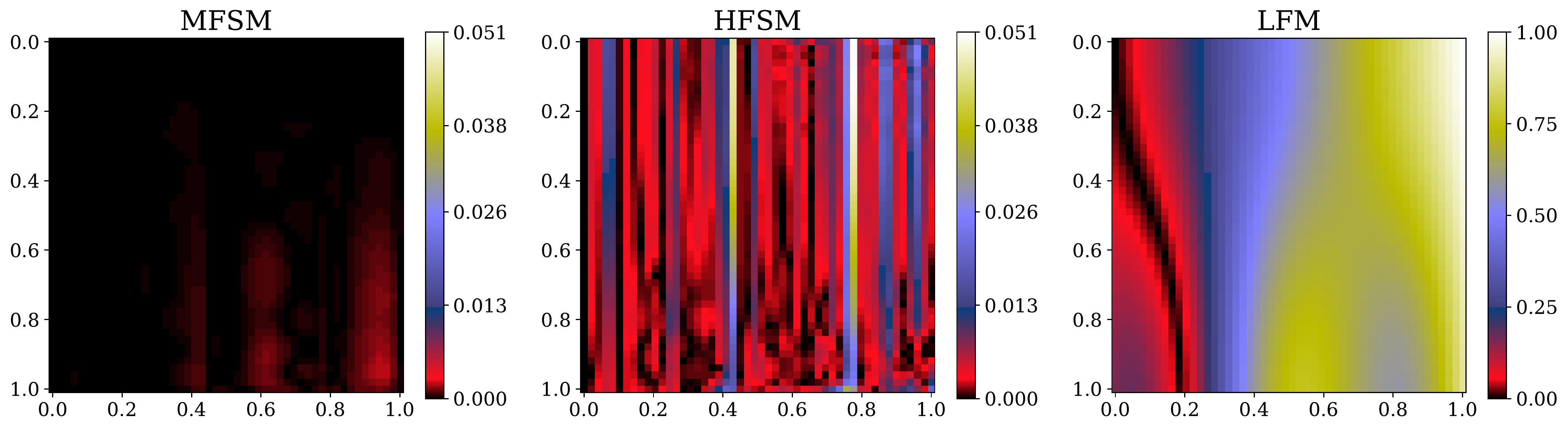}}
    \subfigure[]
    {\label{subfig:lab43}\includegraphics[width=0.9\textwidth]{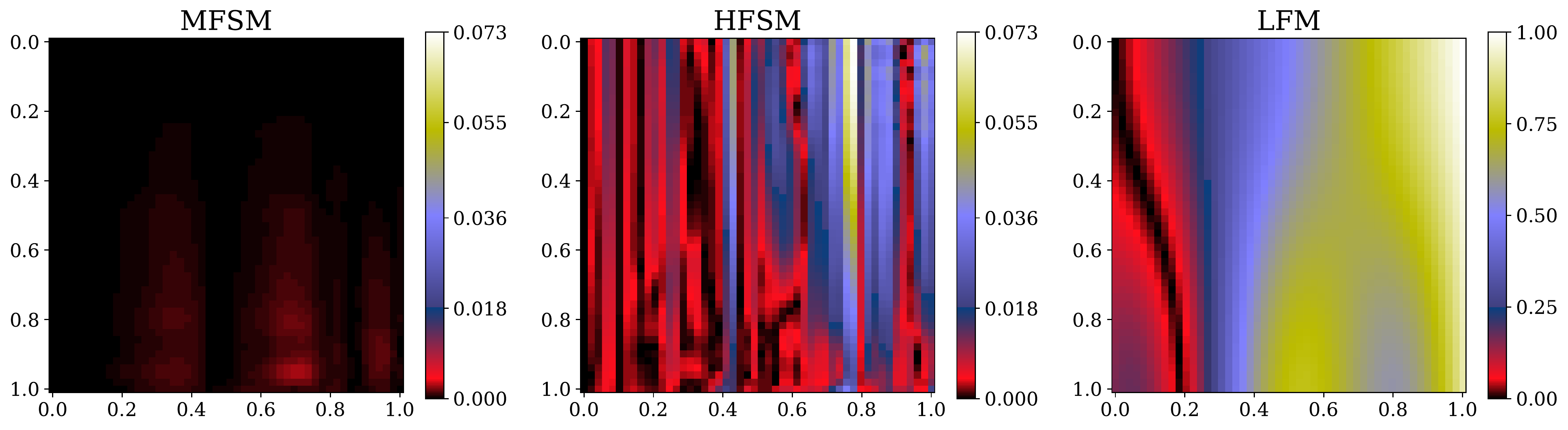}}
    \subfigure[]
    {\label{subfig:lab44}\includegraphics[width=0.9\textwidth]{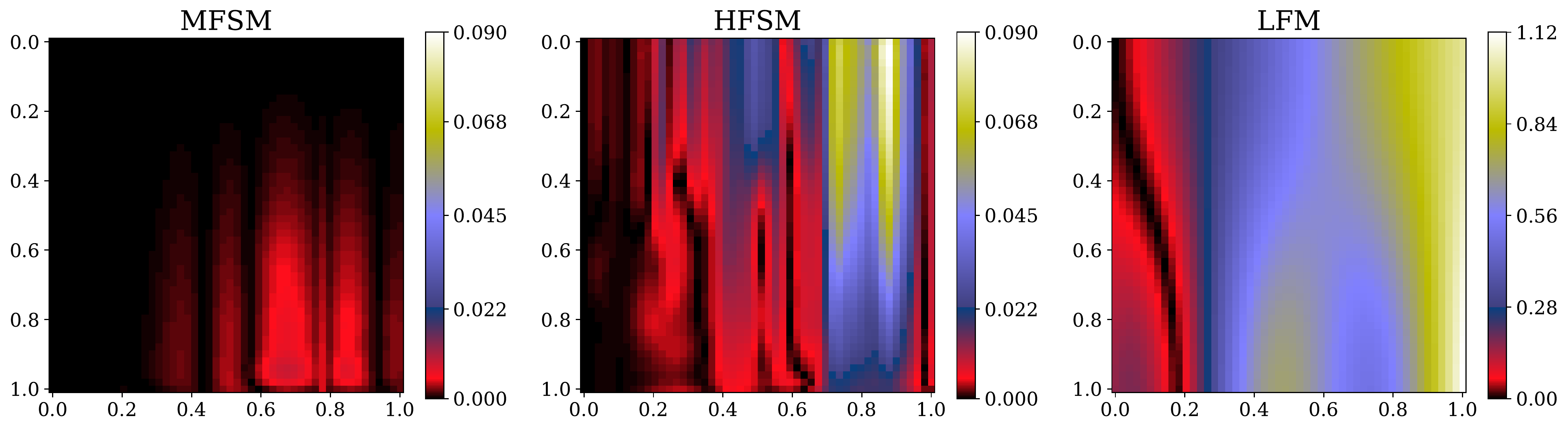}}
    \caption{Absolute error between exact HF solution and prediction from MFSM and HFSM-WNO on an unseen test set along with absolute error between exact HF and LF solution for training dataset sizes of \textbf{(a)} $10$, \textbf{(b)} $8$, \textbf{(c)} $6$, and \textbf{(d)} $2$ for the 2-dimensional problem with nonlinear correlation. The first column shows the absolute error between the exact HF solution and MFSM-WNO prediction, the second column shows the absolute error between the exact HF solution and HFSM-WNO prediction, and the third column is the absolute error between LF and the exact HF solution.}
    \label{fig:5}
\end{figure}

\begin{figure}[htbp!]
    \centering
    \includegraphics[width=0.7\textwidth]{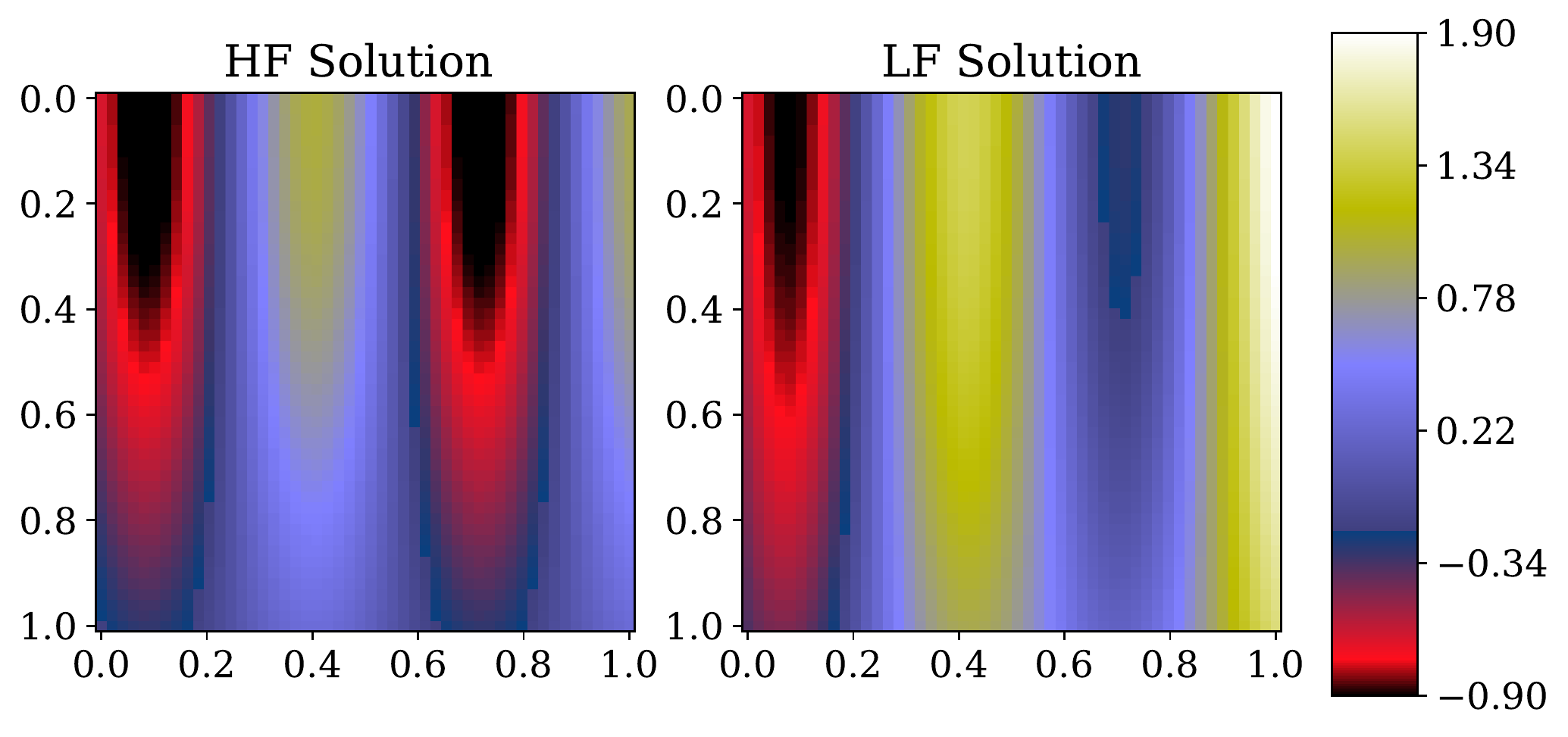}
    \captionof{figure}{Exact HF solution and LF solution for the 2-dimensional problem with nonlinear correlation.}
    \label{fig:4}
\end{figure}

As the next benchmark problem, we subject our network to the learning of complicated nonlinear correlation between HF and LF solutions given by the following equations, 

\begin{equation}
    \begin{aligned}
        D_{L}(a)(x, y) &=\cos (a) \cos (y)+x, \\
        D_{H}(a)(x, y) &=\cos (a) \cos (y)^{2}, \\
        a &=k x-4,
    \end{aligned}
\end{equation}
where $x, y \in[0,1]$ and $k \in[8,10]$. We perform MFSM construction using MF-WNO and DeepONet and HFSM construction using WNO for this case as well. Also, different models are trained on datasets with sizes $2$, $6$, $8$, and $10$, respectively, and the results are presented in Fig. \ref{fig:5} for MFSM-WNO and HF-WNO. In addition, the exact HF solution and LF solution are presented in Fig. \ref{fig:4}. Further, the MSE for MFSM-WNO, MFSM-DeepONet, and HFSM-DeepONet predictions along with $MSE_{\text{LF}}$ are presented in Table \ref{tab:2}.

It can be seen from Table \ref{tab:2} that error for MFSM-WNO predictions is 3 orders of magnitude less than HFSM-WNO predictions for training dataset sizes $10$ and $6$, while for sizes $8$ and $2$ is two orders of magnitude less. Also, predictions from MFSM-DeepONet are closer to those from MFSM-WNO, but it is MFSM-WNO that provides predictions with the least error. Furthermore, the results in Fig. \ref{fig:5} demonstrate the superior performance of MFSM with respect to HFSM for all training dataset sizes. These results have expressed that our model has been successful in discovering the complex nonlinear correlation between the LF and HF solution and has shown the capability of prediction with high orders of accuracy. Furthermore, the inference time for computing the solution for $200$ samples, in this case, is $0.29$ seconds.

\begin{table}[htbp!]
    \centering
    \caption{MSE error between exact HF solution and predictions from MFSM-WNO, HFSM-WNO, and MFSM-DeepONet on unseen test dataset for different training dataset sizes for the 2-dimensional problem with nonlinear correlation.}
    \label{tab:2}
    \begin{tabular}{m{3cm}m{2cm}m{2cm} c} 
    \toprule
    \multirow{2}{*}{$\mathcal{Q}_{train}$ Size} &\multicolumn{3}{c}{MSE}\\ \cline{2-4}
    & MFSM-WNO & HFSM-WNO & MFSM-DeepONet \\ [0.5ex] 
    \hline
    \vspace{0.2em}
    $10$ & 1.2043 $\times 10^{-8}$ & 1.7359 $\times 10^{-5}$ &$6.2326\times 10^{-6}$\\ 
    $8$ & 1.3134 $\times 10^{-7}$ & 7.7837 $\times 10^{-5}$  &$2.7814 \times 10^{-5}$\\ 
    $6$ & 3.5792 $\times 10^{-7}$ & 3.3641 $\times 10^{-4}$  &$5.6104\times 10^{-5}$\\ 
    $2$ & 3.3927 $\times 10^{-5}$ & 2.3504 $\times 10^{-3}$  &$1.2385 \times 10^{-4}$\\ 
    \hline
    $MSE_{LF}$ & \multicolumn{3}{c}{3.2892$\times 10^{-1}$}\\
    \bottomrule
    \end{tabular}
    
\end{table}
\subsubsection{Artificial Benchmark III:  Modified multi-fidelity Forrester function}
\begin{figure}[htbp!]
    \centering
    \subfigure[]
    {\label{subfig:lab051}\includegraphics[width=0.9\textwidth]{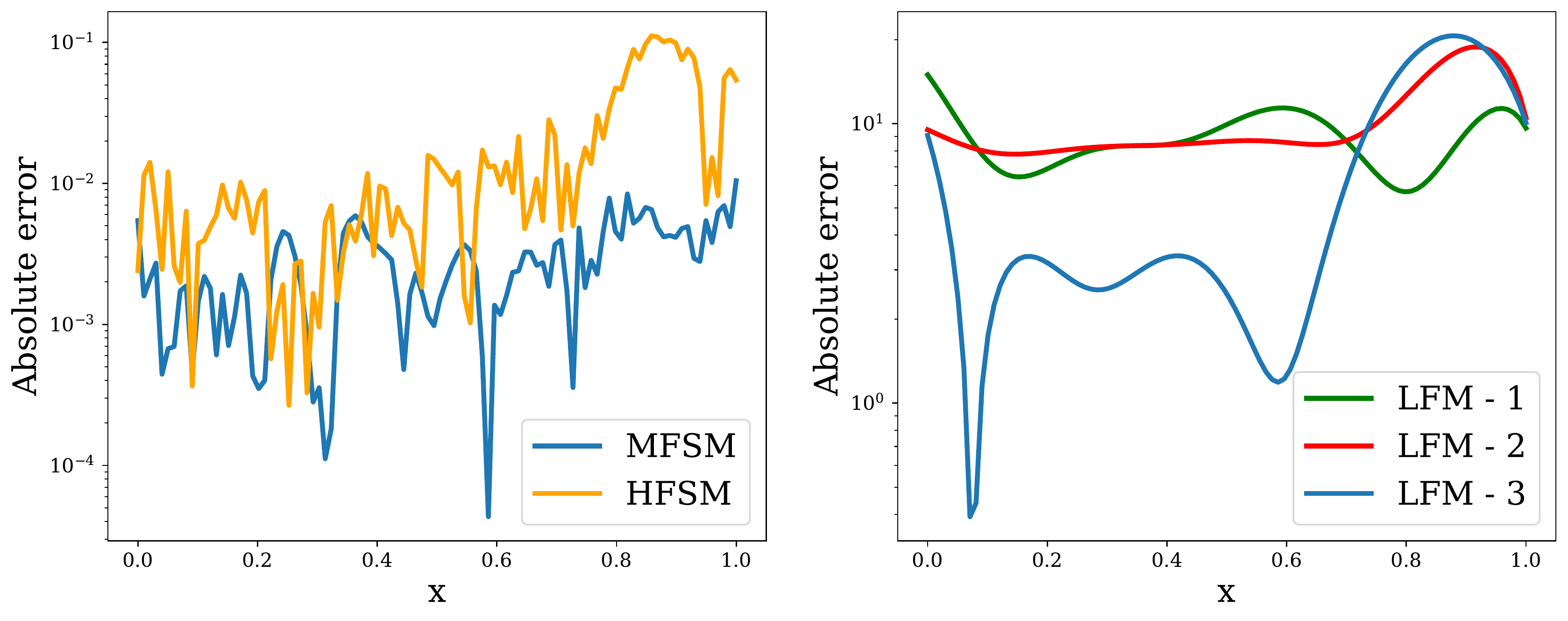}}
    \vspace{2em}
    \subfigure[]
    {\label{subfig:lab055}\includegraphics[width=0.5\textwidth]{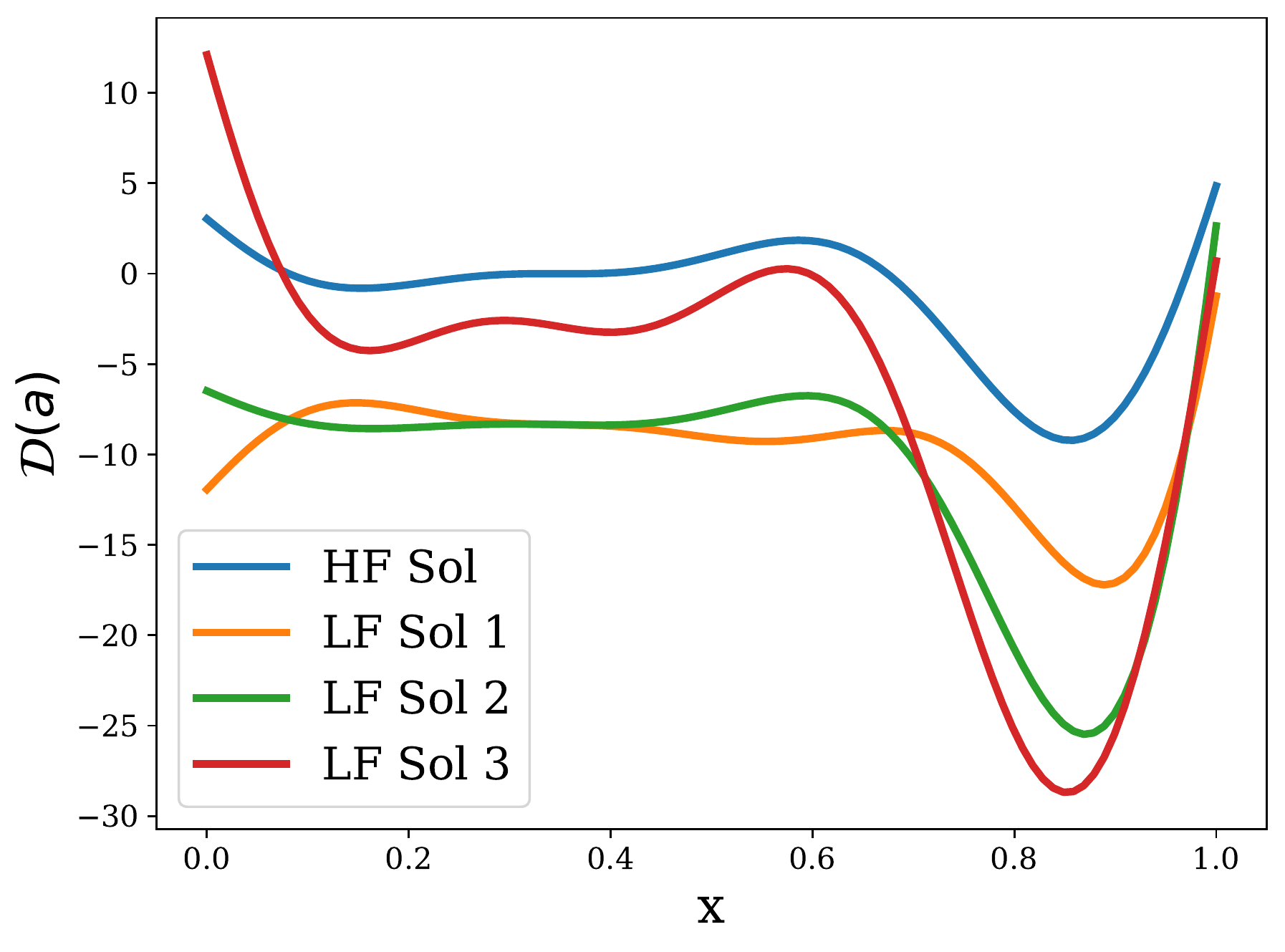}}
    \caption{\textbf{(a)} Absolute error between exact HF solution and prediction from MFSM and HFSM on an unseen test set along with absolute error between exact HF and LF solution for training dataset sizes of  $10$ for modified MF Forester problem. The first column shows the absolute error between the exact HF solution and predictions from MFSM and HFSM, and the second column shows the absolute error between LF solutions and the exact solution HF solution. The y-axis has a logarithmic scale. \textbf{(b)} Plots for the exact HF solution and LF solutions for the modified MF Forester problem.}
    \label{fig:051}
\end{figure}

In order to illustrate the ability of our model to learn from multiple sources, we assess its performance on a modified version of the MF problem put forth by Forrester et al. \cite{forrester2007multi,cheng2021multi}. The considered problem has three LF models which are correlated with the HF model. Mathematically, the considered numerical example is expressed as follows

\begin{equation}
    \begin{aligned}
        D_{L1}(a)(x) &=\left(3 x^2-0.1 x-1.3\right) D_{H}(a)(x)-(x+8), \\
        D_{L2}(a)(x) &=\left(x^3+x^2-0.1 x+0.5\right) D_{H}(a)(x)-(x+8),  \\
        D_{L3}(a)(x) &=(-2 x+4) D_{H}(a)(x)-(x+8) ),  \\
        D_{H}(a)(x) &=(6 x-2)^2 \sin (a), \\
        a &=k x-4,
    \end{aligned}
\end{equation}
where $k\in [10,14]$ and $x \in[0,1]$. Multiple MFSMs and HFSMs for this problem are trained on datasets with different sizes, which are $4$, $6$, $10$, and $15$, respectively, and the obtained results are presented in Fig. \ref{fig:051}. Furthermore, the exact HF solution and the three LF solutions can be found in Fig. \ref{fig:051}. Also, the MSE for MFSM-WNO, MFSM-DeepONet, and HFSM-WNO predictions along with $MSE_{\text{LF}}$ are presented in Table \ref{tab:021}. 
\begin{table}[htbp!]
    \centering
    \caption{MSE error between exact HF solution and predictions from MFSM and HFSM on unseen test dataset for different training dataset sizes for modified MF Forrester problem.}
    \label{tab:021}
    \begin{tabular}{m{3cm}m{2cm}c c }
    \toprule
    \multirow{2}{*}{$\mathcal{Q}_{train}$ Size} &\multicolumn{3}{c}{MSE}\\ \cline{2-4}
    & MFSM-WNO & HFSM-WNO & MFSM-DeepONet \\ [0.5ex] 
    \hline
    \vspace{0.2em}
    $10$ & 2.2198 $\times 10^{-4}$ & 2.9405 $\times 10^{-3}$ & 1.6296 $\times 10^{-1}$\\ 
    $6$ & 7.4916 $\times 10^{-4}$ & 5.5142 $\times 10^{-3}$  & 2.3860 $\times 10^{-1}$\\ 
    $4$ & 2.1632 $\times 10^{-3}$ & 4.0306 $\times 10^{-2}$  & 2.6323\\ 
    $2$ & 8.1014 $\times 10^{-2}$ & 1.7739  & 4.1510\\ 
    \hline
    $MSE_{LF1}$ & \multicolumn{3}{c}{69.1840} \\
    \hline
    $MSE_{LF2}$ & \multicolumn{3}{c}{80.8175} \\
    \hline
    $MSE_{LF3}$ & \multicolumn{3}{c}{54.3353} \\
    \bottomrule
    \end{tabular}
    
\end{table}
It is evident from the results that MFSM-WNO has good predictive capabilities in comparison to the baselines, which is a testament to the abilities of the developed framework to handle data from multiple sources. Finally, we evaluate that MFSM-WNO takes $0.020$ seconds for predicting the solution for $200$ samples for this problem.
\subsubsection{Artificial Benchmark IV:  Multi-fidelity example with a nonlinear bias}
\begin{figure}[htbp!]
    \centering
    \subfigure[]{\label{subfig:lab061}\hspace{-2.5em}\includegraphics[width=0.41\textwidth]{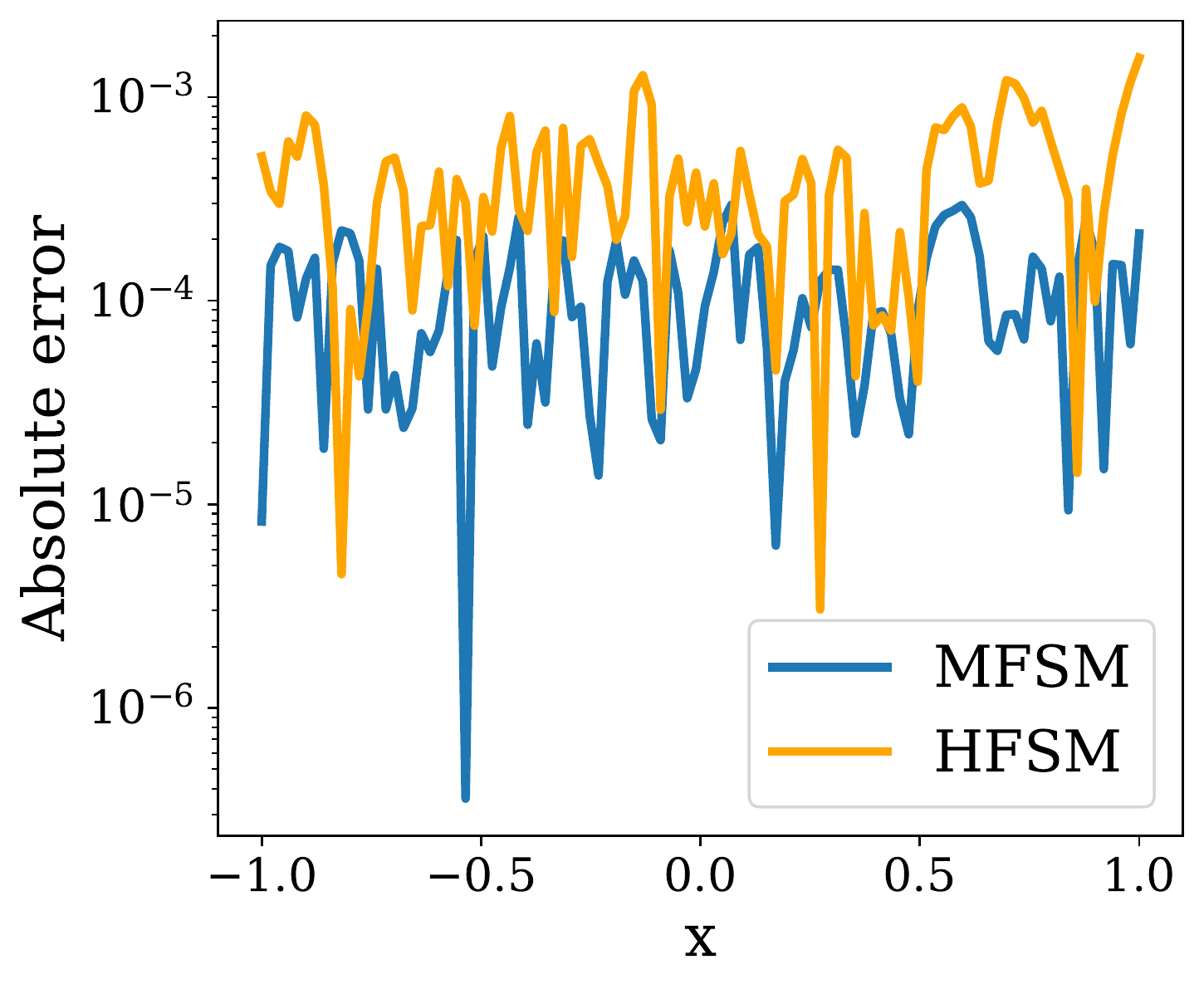}}
    \hspace*{2em}
    \subfigure[]{\label{subfig:lab065}\hspace{-1.4em}\includegraphics[width=0.41\textwidth]{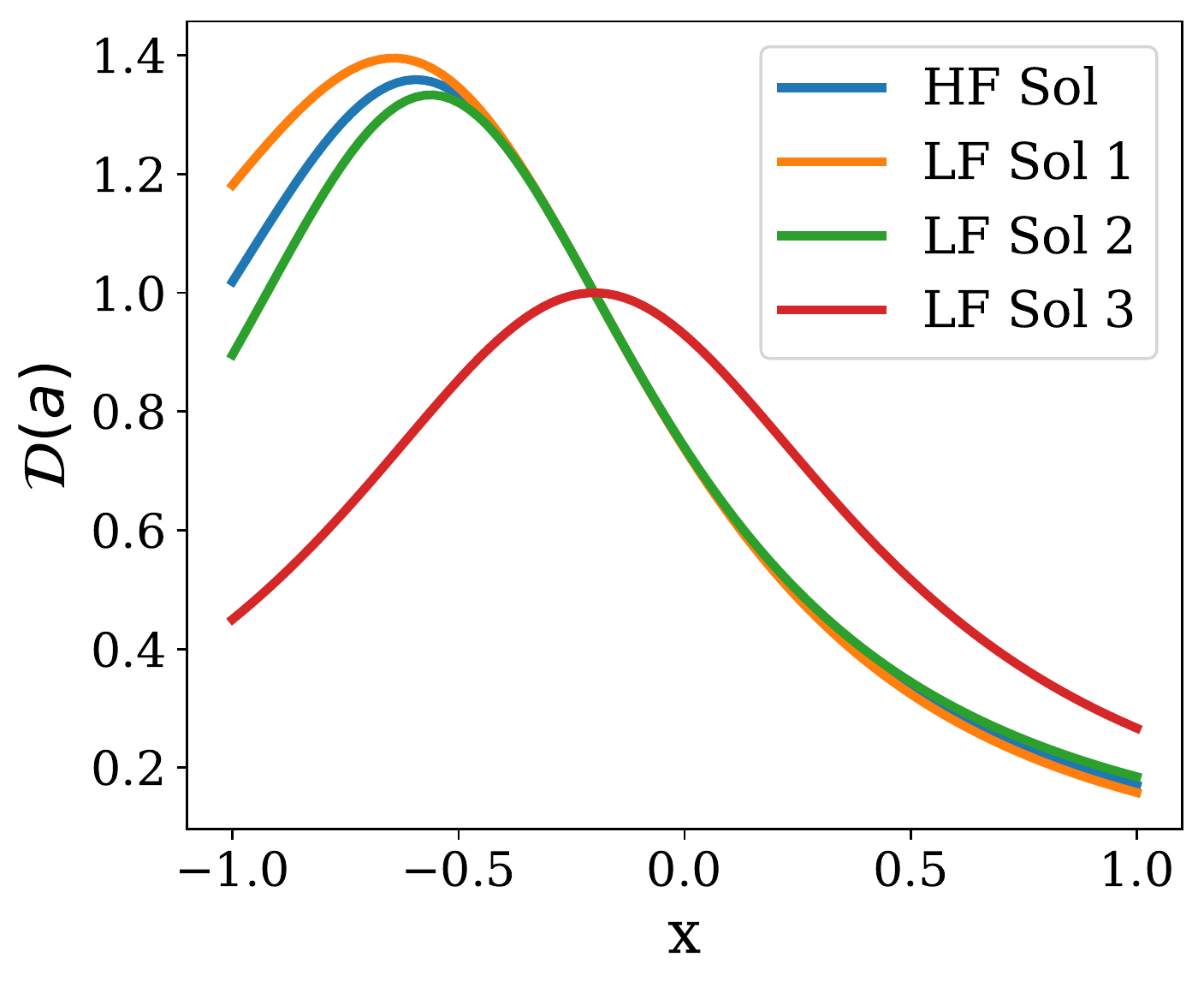}}
    \caption{\textbf{(a)} Absolute error between exact HF solution and prediction from MFSM-WNO and HFSM-WNO on an unseen test set for training dataset sizes of  $10$ for MF example with nonlinear bias. The y-axis has a logarithmic scale. \textbf{(b)} Plot for the exact HF solution and LF solutions for the MF problem with nonlinear bias.}
    \label{fig:061}
\end{figure}

The developed framework is agnostic to the relationship between different fidelities. Also, it has strong nonlinear operator approximation abilities. To test these abilities, we subject our model to the estimation of a correlation where the bias is nonlinear \cite{eweis2022data}. The example problem can be represented as follows:

\begin{equation}
    \begin{aligned}
        D_{L1}(a)(x) &= \dfrac{1}{0.2a^3+ a^2 + a +1}, \\
        D_{L2}(a)(x) &= \dfrac{1}{a^2 + a +1},  \\
        D_{L3}(a)(x) &= \dfrac{1}{a^2 +1},  \\
        D_{H}(a)(x) &= \dfrac{1}{0.1a^3+ a^2 + a +1}, \\
        a &=k x,
    \end{aligned}
\end{equation}  
where $k\in [0.1,1.2]$ and $x \in[-2,3]$. The MFSM-WNO, MFSM-DeepONet, and HFSM-WNO are constructed using a training dataset of size $10$. The obtained results are presented in Table \ref{tab:022}, which indicate the superiority of MFSM-WNO over the models in contention. Furthermore, an illustration of absolute errors for HFSM-WNO and MFSM-WNO predictions along with the exact HF and LF solutions are presented in Figure \ref{fig:061}. Furthermore, the time taken by MFSM-WNO for predicting the solution for $200$ samples is $0.014$ seconds.
\begin{table}[htbp!]
    \centering
    \caption{MSE error between exact HF solution and predictions from MFSM-WNO, HFSM-WNO, and MFSM-DeepONet trained using a dataset of size $10$ on unseen test dataset for MF example with nonlinear bias.}
    \label{tab:022}
    \begin{tabular}{m{3cm}m{2cm}c c }
    \toprule
    \multirow{2}{*}{$\mathcal{Q}_{train}$ Size} &\multicolumn{3}{c}{MSE}\\ \cline{2-4}
    & MFSM-WNO & HFSM-WNO & MFSM-DeepONet \\ [0.5ex] 
    \hline
    \vspace{0.2em}
    $10$ & 5.7937 $\times 10^{-7}$ & 1.4691 $\times 10^{-5}$ & 9.5735 $\times 10^{-3}$\\ 
    \bottomrule
    \end{tabular}
    
\end{table}

\subsubsection{Artificial Benchmark V:  High-dimensional bi-fidelity example}
\begin{figure}[htbp!]
    \centering
    \subfigure[]{\label{subfig:lab071}\hspace{-2.5em}\includegraphics[width=0.46\textwidth]{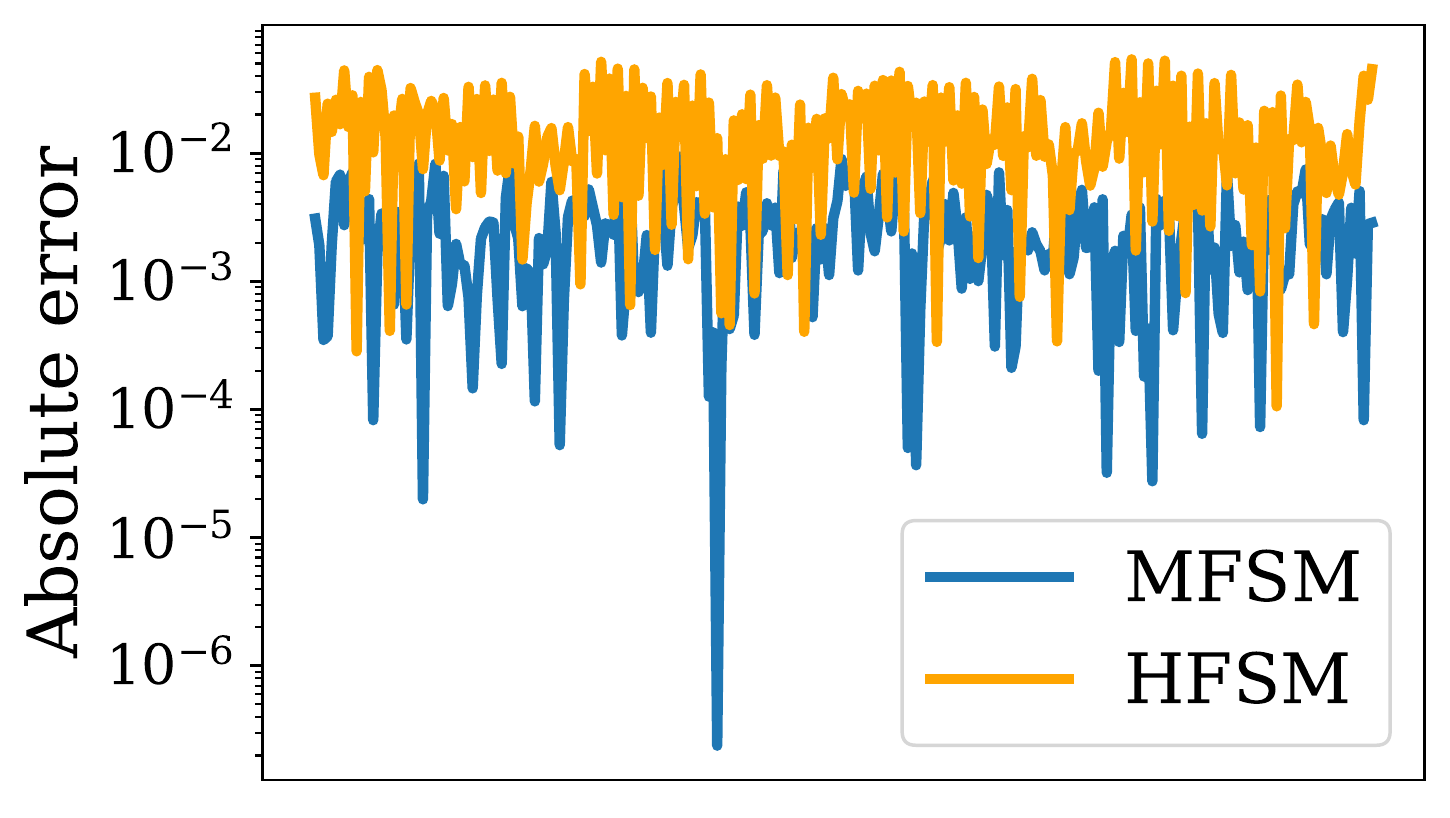}}
    \hspace*{2em}
    \subfigure[]{\label{subfig:lab075}\hspace{-1.4em}\includegraphics[width=0.4\textwidth]{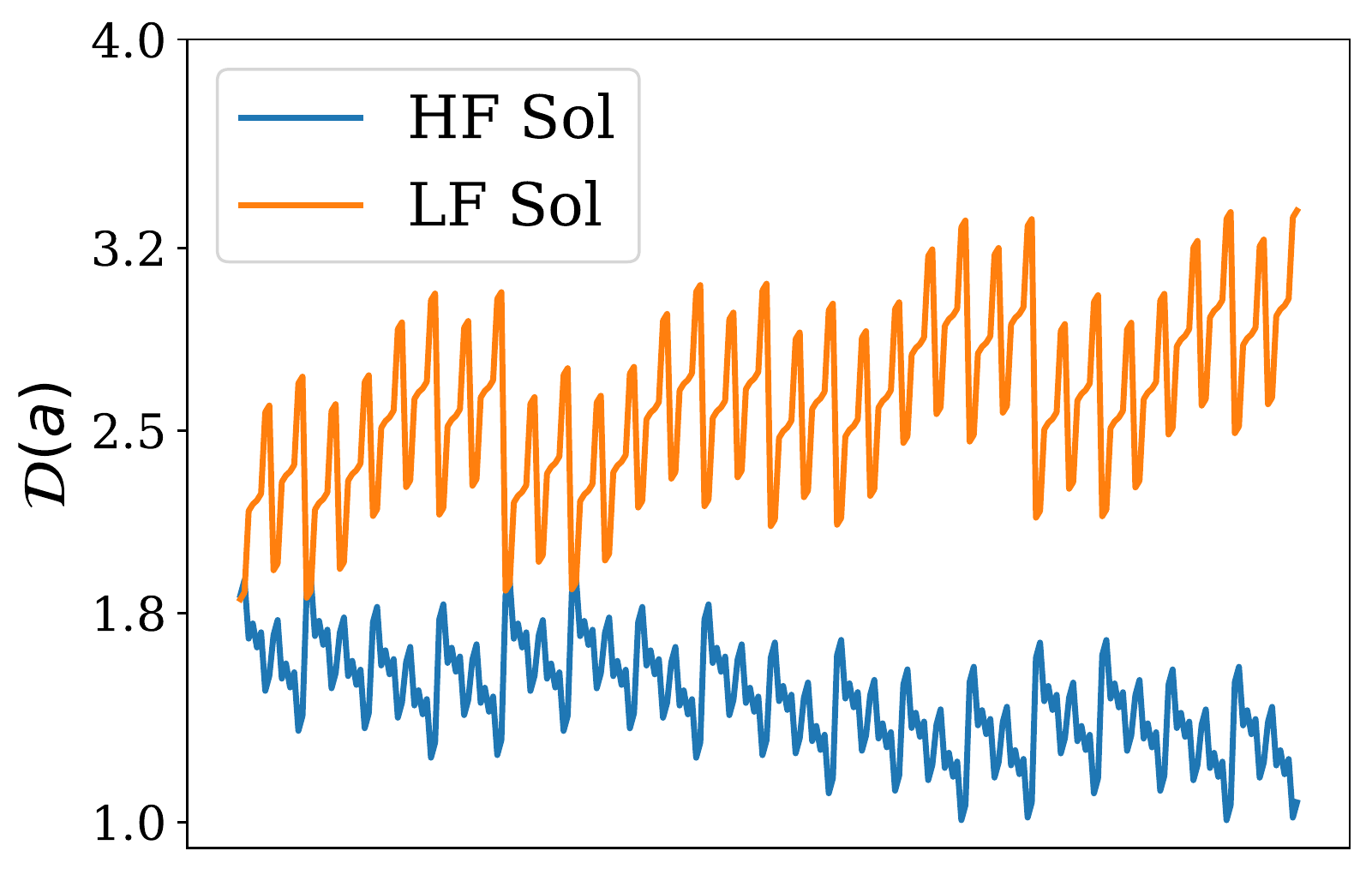}}
    \caption{\textbf{(a)} Absolute error between exact HF solution and prediction from MFSM-WNO and HFSM-WNO on an unseen test set for training dataset sizes of  $10$ for eight dimension bi-fidelity problem. The y-axis has a logarithmic scale. \textbf{(b)} Plot for the exact HF solution and LF solution for the eight-dimensional bi-fidelity problem.}
    \label{fig:071}
\end{figure}

In this section, in order to assess the ability of MF-WNO in tackling problems in higher dimensions, we consider the following eight-dimensional MF example problem,
\begin{equation}
    \begin{aligned}
        D_{L}(a)(x) &= -2.4+\sum_{i=1}^8\left(\sin a_i+\sin ^2\left(\frac{16 a_i}{15}-1\right)\right), \\
        D_{H}(a)(x) &= 2.4+\sum_{i=1}^8\left[\sin \left(\frac{16 a_i}{15}-1\right)+\sin ^2\left(\frac{16 a_i}{15}-1\right)\right], \\
        a_i &=k x_i,
    \end{aligned}
\end{equation}
where $k\in[0.8,1.2]$ and $x_i\in [-1,1]$. We test the ability of MF-WNO by creating a surrogate using only $10$ training samples in the training process. Simultaneously, HFSM and MF-DeepONet are also trained using  $10$ training samples. Also, a one-dimensional WNO is used for both HFSM and MFSM in this case. The results are presented in Table \ref{tab:023}. Also, a pictorial illustration of absolute errors with respect to exact solution and exact HF and LF solutions are presented in Figure \ref{fig:071}. Figure \ref{subfig:lab075} makes it apparent that the desired solution is very irregular with a large number of spikes or discontinuities. Furthermore, Table \ref{tab:023} indicates that both MFSM-DeepONet and MFSM-WNO are able to outperform HFSM-WNO. Regardless, it is MFSM-WNO that has the best performance. Also, the prediction time for $200$ samples is $0.021$ seconds. 

\begin{table}[htbp!]
    \centering
    \caption{MSE error between exact HF solution and predictions from MFSM-WNO, HFSM-WNO, and MFSM-DeepONet trained using a dataset of size $10$ on an unseen test dataset for a training dataset size of $10$ for eight-dimensional MF problem.}
    \label{tab:023}
    \begin{tabular}{m{3cm}m{2cm}c c }
    \toprule
    \multirow{2}{*}{$\mathcal{Q}_{train}$ Size} &\multicolumn{3}{c}{MSE}\\ \cline{2-4}
    & MFSM-WNO & HFSM-WNO & MFSM-DeepONet \\ [0.5ex] 
    \midrule
    \vspace{0.2em}
    $10$ & 2.1557 $\times 10^{-5}$ & 2.2795 $\times 10^{-4}$ & 9.02378$\times 10^{-5}$\\ 
    \bottomrule
    \end{tabular}
    
\end{table}

\subsection{Problem Set II: Stochastic Poisson's equation and Darcy flow in an irregular domain }
\subsubsection{Stochastic Poisson's equation}
\begin{figure}[htbp!]
\centering
\subfigure[]{\label{subfig:lab61}\includegraphics[width=0.49\textwidth]{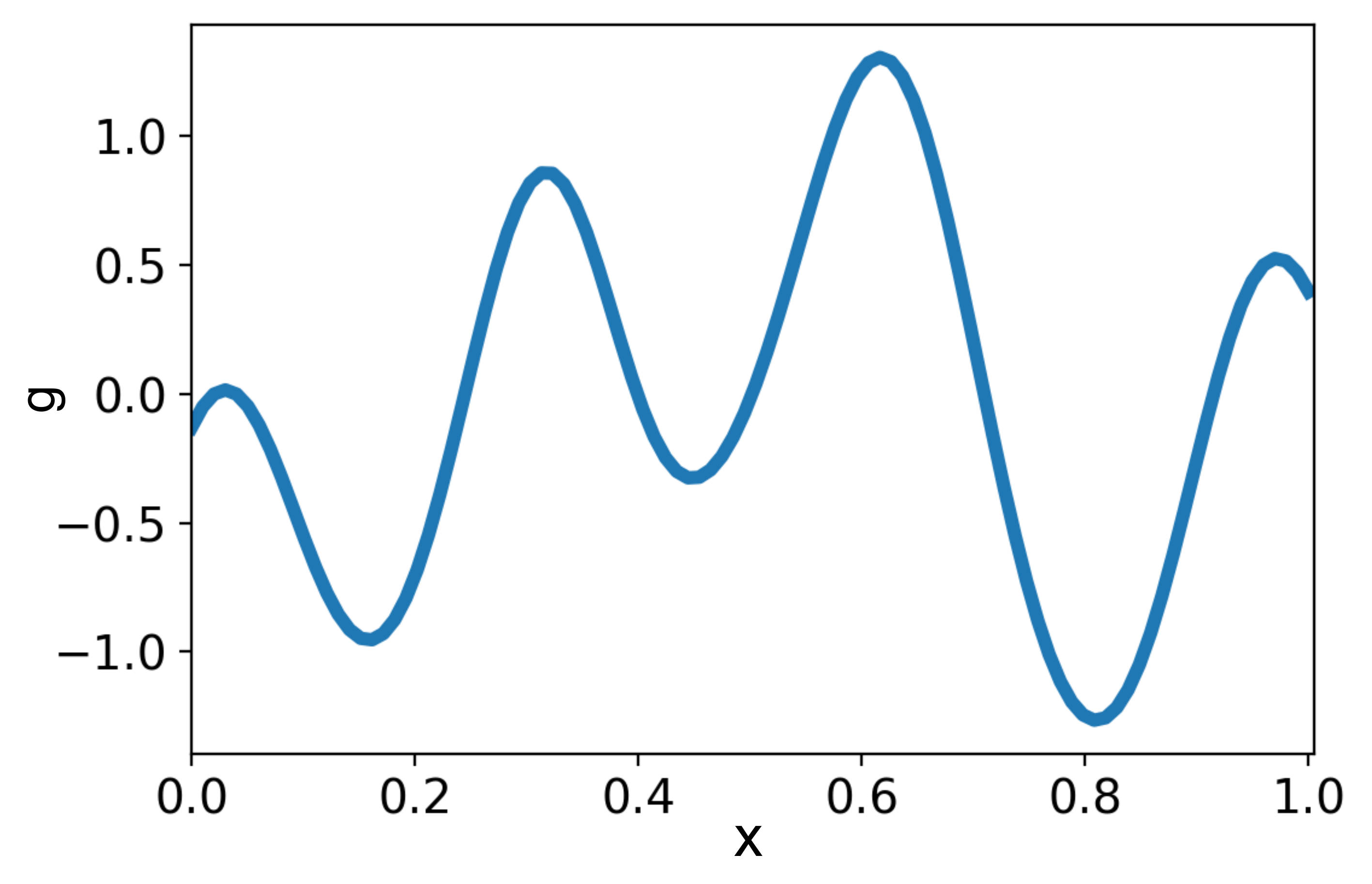}}
\subfigure[]{\label{subfig:lab62}\includegraphics[width=0.49\textwidth]{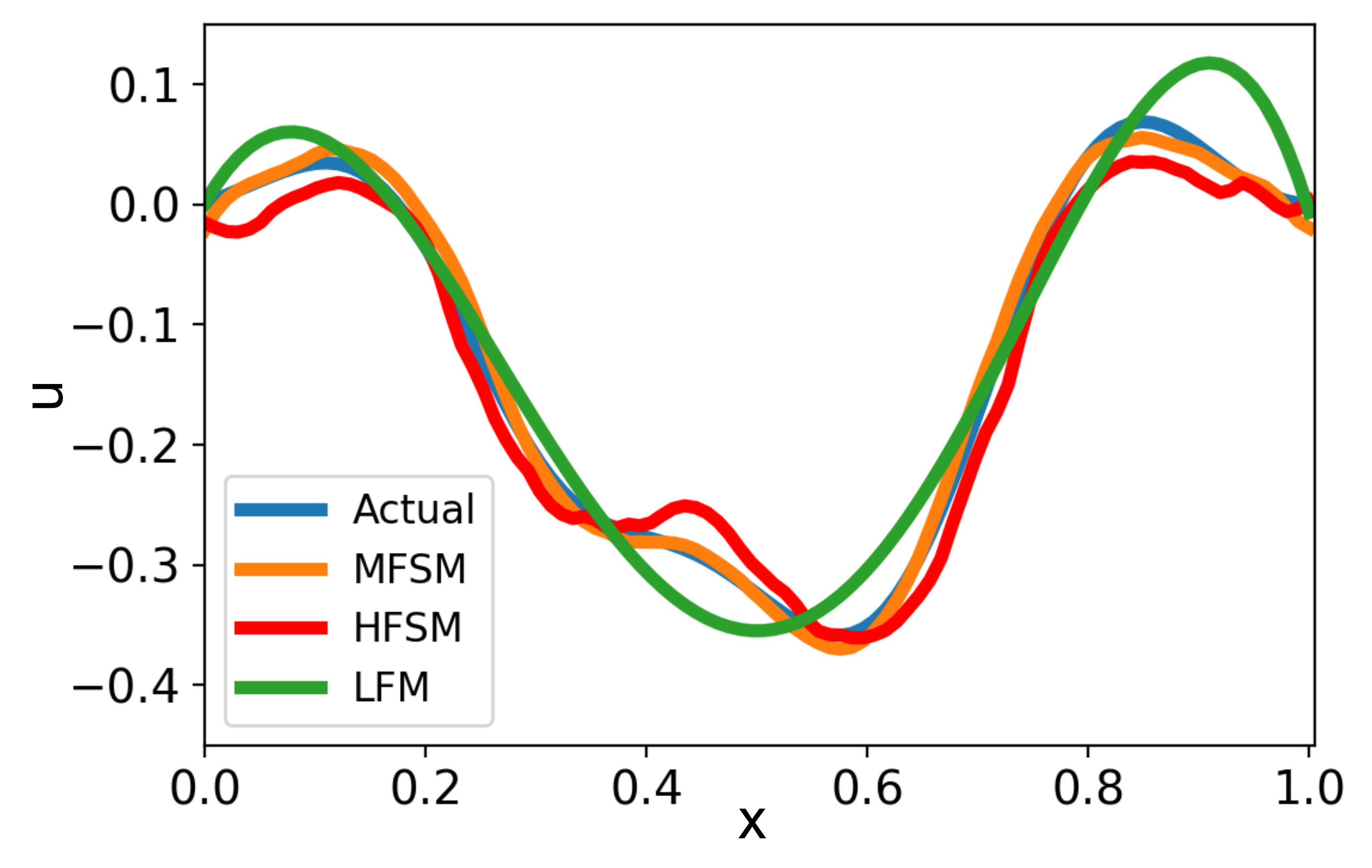}}
\caption{\textbf{(a)} A randomly sampled instance of the stochastic forcing term $g(x)$.
\textbf{(b)} Comparison of exact HF solution with the solution predicted by MFSM-WNO, HFSM-WNO, and low-fidelity model (LFM)(or solver, in the current case) for a single test instance for models trained with dataset size $25$ for stochastic Poisson's equation.}
\label{fig:6}
\end{figure}

The following ordinary differential equation (ODE) is known as the 1-dimensional Poisson's equation,

\begin{equation}
    \frac{d^{2} u}{d x^{2}}=20 g(x), \quad x \in[0,1], \quad u(0)=u(1)=0,
\end{equation}
where $g(x)$ is the spatially varying forcing term. In the current example, we model the forcing term as Gaussian random field (GRF) as,

\begin{equation}
    g(x) \sim \mathcal{G} \mathcal{P}\left(m(x), k\left(x, x^{\prime}\right)\right),
\end{equation}
where the mean $m(x)$ is zero and the covariance function is modeled using a Gaussian kernel as, $k\left(x, x^{\prime}\right)=\exp \left(-(x-x^{\prime})^{2} /\left(2 l^{2}\right)\right)$. Furthermore, the parameter for lengthscale $l = 0.1$. The goal of this problem is to learn the operator mapping from the stochastic forcing term to the solution of the ODE:

\begin{equation}
    \mathcal{D}: g(x) \mapsto u.
\end{equation}
An instance of $g(x)$ which has been randomly sampled is provided in Fig. \ref{fig:6}. The solver for Poisson's equation is based on the finite difference method. The HF and LF dataset, in this case, is generated using a finer and coarser grid. The mesh size $\Delta x$ for the fine grid is $1/99$, while for the coarse grid, it is $1/9$. The finite difference solver used for data generation is based on the code provided by Lu et al. \cite{lu2022multifidelity}. For this example as well, we train different MFSMs and HFSMs using training datasets with sizes of $25$, $20$, $10$, and $5$. Furthermore, we also present the MSE value for MFSM and HFSM predictions for all training dataset sizes in Table \ref{tab:3}.\par

\begin{table}[htbp!]
    \centering
    \caption{MSE error between exact HF solution and predictions from MFSM-WNO, HFSM-WNO, and MFSM-DeepONet on unseen test dataset for different training dataset sizes for stochastic Poisson's equation.}
    \label{tab:3}
    \begin{tabular}{m{3cm}m{2cm}m{2cm} c} 
    \toprule
    \multirow{2}{*}{$\mathcal{Q}_{train}$ Size} &\multicolumn{3}{c}{MSE}\\ \cline{2-4}
    & MFSM-WNO & HFSM-WNO & MFSM-DeepONet \\ [0.5ex] 
    \midrule
    \vspace{0.2em}
    $25$ & 3.2329 $\times 10^{-4}$ & 2.5602 $\times 10^{-3}$ & 1.6757 $\times 10^{-3}$\\ 
    $20$ & 5.1759 $\times 10^{-4}$ & 6.3312 $\times 10^{-3}$ & 1.9567 $\times 10^{-3}$\\ 
    $10$ & 7.8373 $\times 10^{-4}$ & 1.8202 $\times 10^{-2}$ & 2.3586 $\times 10^{-3}$\\ 
    $5$ & 1.4672 $\times 10^{-3}$ & 1.3396 $\times 10^{-1}$  & 2.6248  $\times 10^{-3}$\\ 
    \hline
    $MSE_{LF}$ & \multicolumn{3}{c}{2.0642 $\times 10^{-3}$}\\
    \bottomrule
    \end{tabular}
    
\end{table}
It is clear from Table \ref{tab:3} that MSE values for MFSM-WNO test predictions are two orders less in magnitude as compared to HFSM-WNO's prediction for training dataset sizes of $10$ and $4$, while a decrease in error by one order of magnitude is found for sizes $25$ and $20$. Furthermore, overall, the MSE value for MFSM-WNO prediction is better than the output from the LF solver and MFSM-DeepONet. Also, the performance for a single case test case can be assessed from Figure \ref{fig:6}. At last, the time taken for predicting the solution for $200$ samples by MFSM-WNO is $0.021$ seconds.

\subsubsection{2-dimensional Darcy flow in a triangular domain with a notch}\label{ss:33}
\begin{figure}[ht!]
    \centering
    \subfigure[]{\label{subfig:lab71}\includegraphics[width=0.4\textwidth]{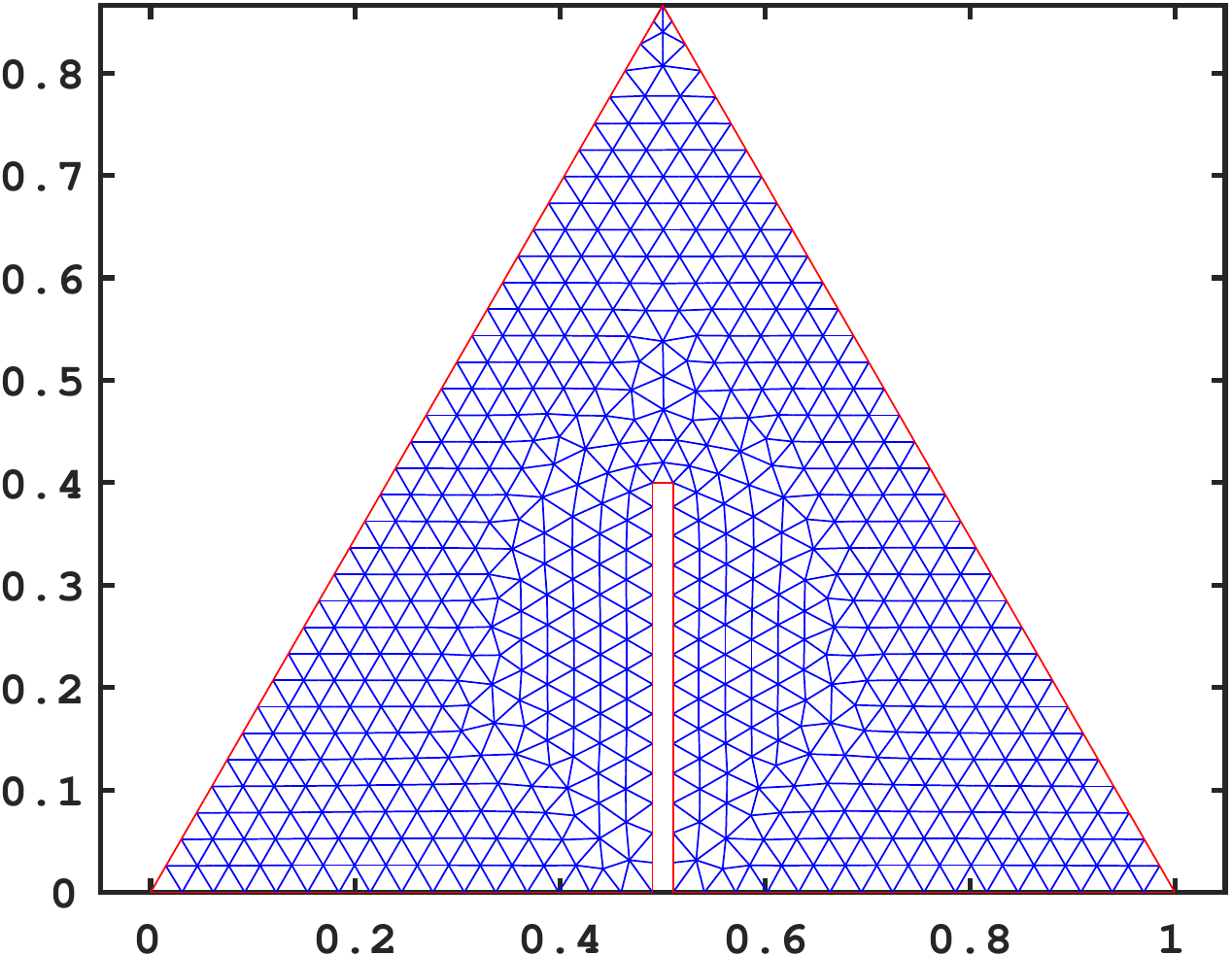}}
    \subfigure[]{\label{subfig:lab72}\includegraphics[width=0.4\textwidth]{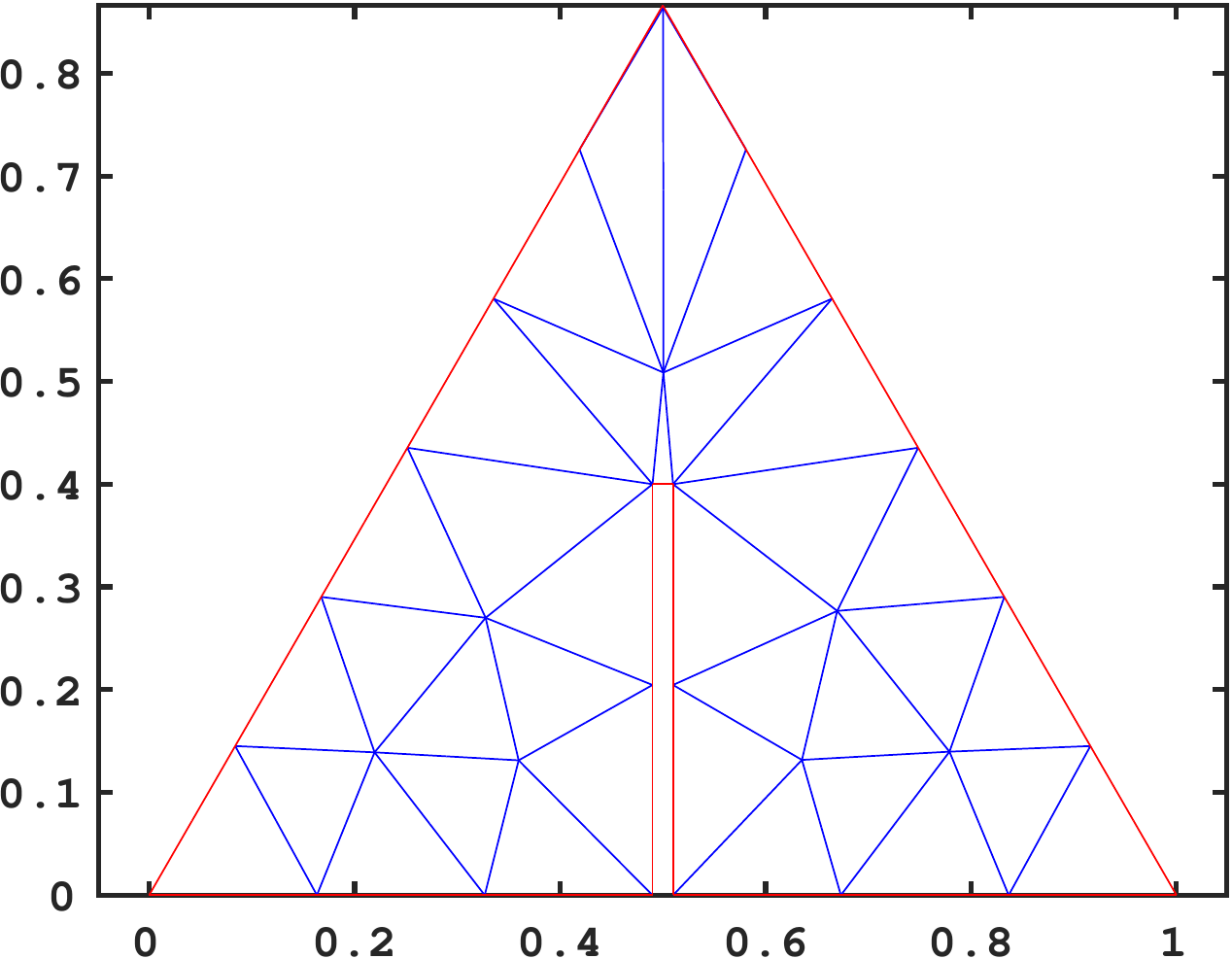}}
    \caption{Comparison of grids used for obtaining LF and HF dataset. \textbf{(a)} Fine grid
    \textbf{(b)} Coarse grid}
    \label{fig:7}
    \end{figure}

For this example, we consider a 2-dimensional elliptic Darcy flow PDE given as,

\begin{equation}
    -\nabla \cdot(a(x, y) \nabla u(x, y))=f(x, y) ; \;\;x, y \in(0, \mathbb{R}),
\end{equation}

where $a(x, y)$ is the permeability field, $u(x, y)$ is the pressure, and $f(x, y)$ is the source term. The Darcy flow equation has been widely used to model the flow through porous media, and it finds substantial applications in fields such as geotechnical, civil, and petroleum engineering. For the current example, a triangular domain with a rectangular notch is considered. The boundary conditions $u(x)\mid_{\partial \omega}$ for this problem are generated using Gaussian process, $\mathrm{GP}\left(0, \mathcal{K}\left(x, x^{\prime}\right)\right)$. Furthermore, the covariance kernel $\mathcal{K}\left(x, x^{\prime}\right)$ is modeled as follows:

\begin{equation}
      \mathcal{K}\left(x, x^{\prime}\right)=\exp \left(-\left(x-x^{\prime}\right)^{2} / 2 l^{2}\right); \;\;x, x^{\prime} \in[0,1].
\end{equation}
Here, the kernel length scale parameter $l$ is set equal $0.2$. Furthermore, $a(x,y) = 0.1$ and $f(x,y) = -1$. Our objective with this problem is to learn the operator mapping from boundary conditions to the pressure field in the whole domain. This could be mathematically expressed as,

\begin{equation}
    \mathcal{D}: u(x, y) \mid_{\partial \omega} \mapsto u(x, y).
\end{equation}
The training data is generated using MATLAB PDE Toolbox and by modification of the code provided by Lu et al. \cite{lu2022comprehensive}. The LF and HF datasets are generated using coarse and finer grids. A pictorial representation of the coarse and fine grids is provided in Fig. \ref{fig:7}. Furthermore, HFSMs and MFSMs are trained using datasets with sizes $25$, $20$, $15$, and $10$. The MSE values for predictions made by the models trained using different dataset sizes on an unseen test dataset, i.e., boundary conditions that are different from what the model was trained on, are provided in Table \ref{tab:4}.

    \begin{figure}[htbp!]
    \centering
    \includegraphics[width=0.8\textwidth]{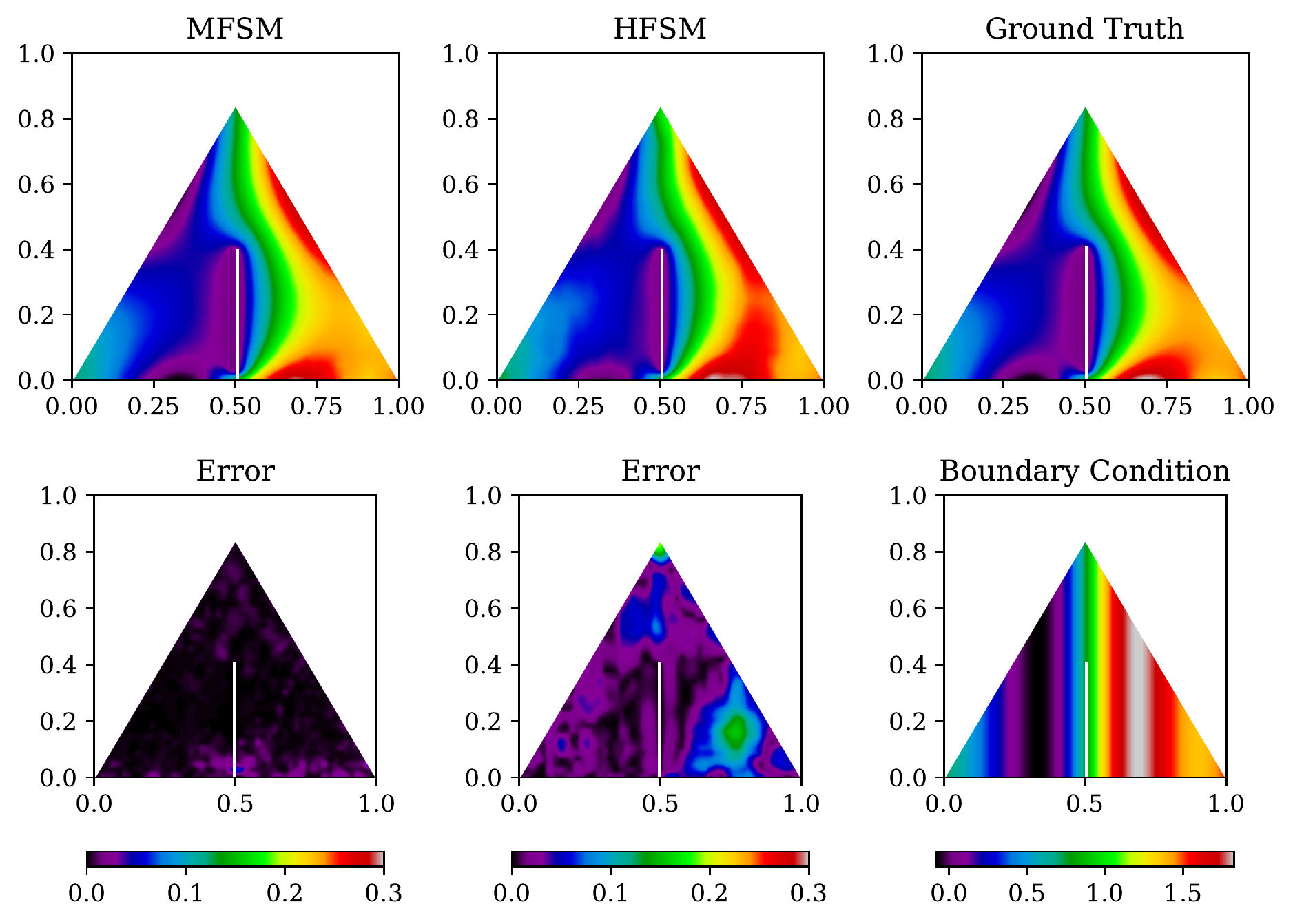}
    \caption{ Comparison of exact HF solution with the solution predicted by MFSM-WNO and predictions along with respective absolute errors and pictorial representation of boundary conditions for a single test instance for models trained with dataset size $30$ for Darcy flow in an irregular domain.}
    \label{fig:8}
\end{figure} 

\begin{table}[htbp!]
    \centering
    \caption{MSE error between exact HF solution and predictions from MFSM-WNO, HFSM-WNO, and MFSM-DeepONet on unseen test dataset for different training dataset sizes for Darcy flow in an irregular domain.}
    \label{tab:4}
    \begin{tabular}{m{3cm}m{2cm}m{2cm} c} 
    \toprule
    \multirow{2}{*}{$\mathcal{Q}_{train}$ Size}&\multicolumn{3}{c}{MSE}\\ \cline{2-4}
    & MFSM-WNO & HFSM-WNO & MFSM-DeepONet \\ [0.5ex] 
    \midrule
    \vspace{0.2em}
    $30$ & 7.8612 $\times 10^{-5}$ & 4.7026 $\times 10^{-3}$ & 1.8963 $\times 10^{-3}$\\ 
    $20$ & 1.2727 $\times 10^{-4}$ & 8.4175 $\times 10^{-3}$ & 2.2284 $\times 10^{-3}$\\ 
    $15$ & 2.7316 $\times 10^{-4}$ & 1.9218 $\times 10^{-2}$ & 2.3136 $\times 10^{-3}$\\ 
    $10$ & 5.6001 $\times 10^{-4}$ & 6.2372 $\times 10^{-2}$ & 2.5950 $\times 10^{-3}$\\ 
    \hline
    $MSE_{LF}$ & \multicolumn{3}{c}{2.6884 $\times 10^{-3}$}\\
    \bottomrule
    \end{tabular}
    
\end{table}
Table \ref{tab:4} reveals that predictions made by MFSM-WNO for all training dataset sizes on an unseen test dataset, in general, have an MSE value that is a couple of orders of magnitude less than the predictions from HFSM-WNO and one order of magnitude less than MFSM-DeepONet. It is also revealed by Table \ref{tab:4} that the MFSM-WNO provides an improvement in error values by two orders of magnitude for training dataset size $30$ and one order improvement for the rest of the dataset sizes over the LF solutions. Furthermore, the outstanding prediction capabilities are made indisputable by the pictorial representation of the comparison of MFSM-WNO and HFSM-WNO predictions with exact HF solution or the ground truth in Fig. \ref{fig:8}. Moreover, the prediction time for $200$ samples, in this case, is $0.576$ seconds.
\subsubsection{Behaviour of single fidelity HFSM-WNO with increasing training samples and comparison with MFSM-WNO}
\begin{figure}[ht!]
    \centering
    \includegraphics[width=0.6\textwidth]{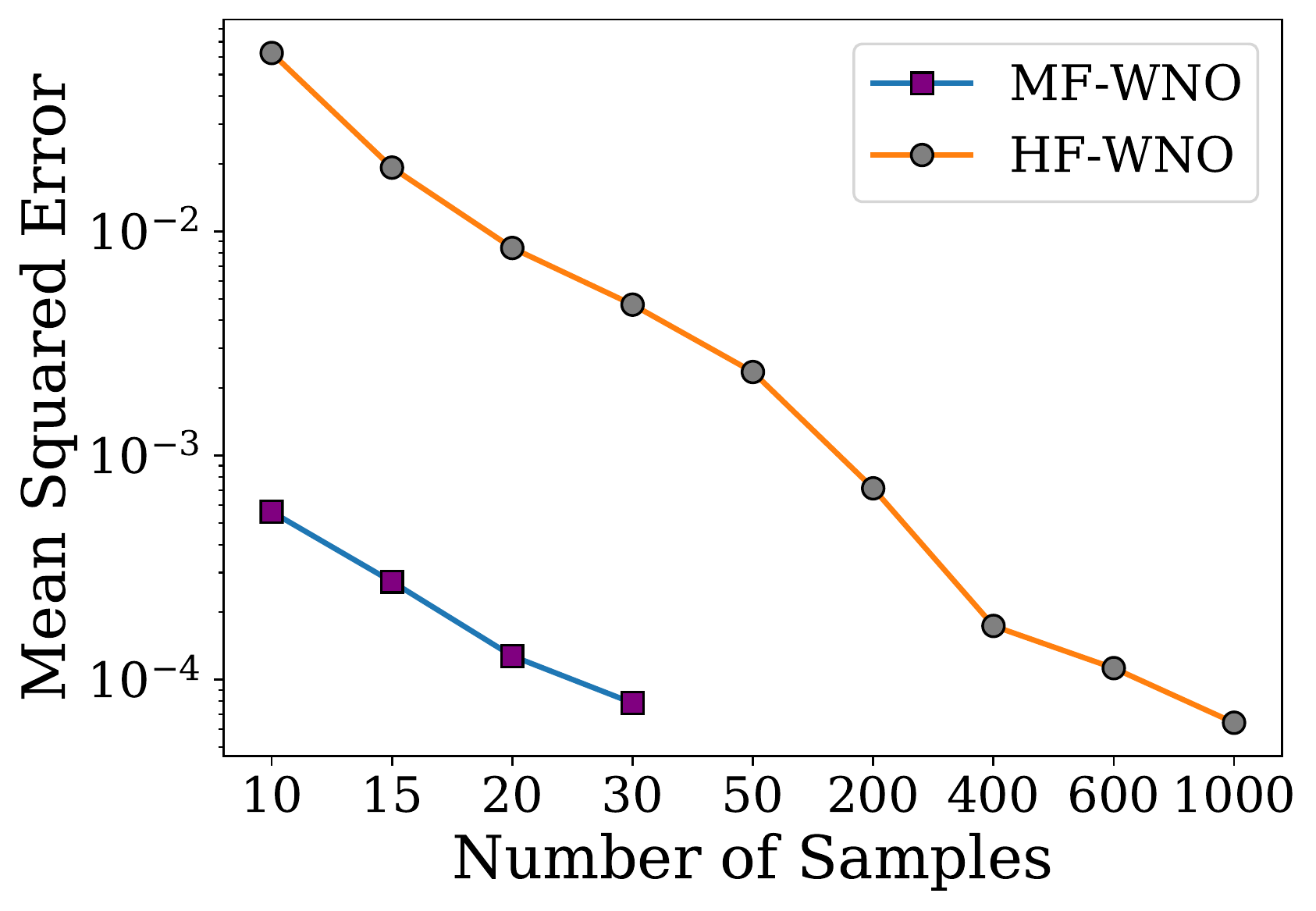}
    \caption{Plot of MSE for HFSM-WNO with respect to increasing training sample number and comparison with MFSM-WNO for the Darcy flow case in an irregular domain.}
    \label{fig:09}
\end{figure}

Our goal with this section is to assess how much advantage our MFSM-WNO provides over the single fidelity HFSM-WNO. The problem, 2-dimensional Darcy flow in a triangular domain with a notch, is considered as the test bed for this analysis. The analysis is conducted by increasing the number of training samples for HFSM-WNO and determining the number of training samples at which the HFSM-WNO obtains similar accuracy compared to our MFSM-WNO. The results containing the pictorial depiction of MSE for increasing training samples are presented in Figure \ref{fig:09}
It can be seen from Figure \ref{fig:09} that to achieve an accuracy similar to MFSM trained using $30$ training samples on an unseen test set (the same test set is used for MFSM and HFSM), the HFSM has to be trained with training samples upwards of $600$ samples. This illustrates the need for good MFSMs and also demonstrates the satisfactory performance of the developed framework in the low data limit.
\subsection{Problem Set III: 2-dimensional stochastic steady state heat equation with application to uncertainty quantification}
As our final problem, we consider the following 2-dimensional elliptic PDE given as,

\begin{equation}
    -\nabla\cdot(a(x,y) \nabla u(x,y))=0;\quad x, y \in(0, \mathbb{R}),
\end{equation}
where $a(x,y)$ is the diffusion coefficient. The equation is defined on a unit square domain with $x \times y \in(0,1)^{2}$ and is subjected to the following boundary conditions:

\begin{equation}
    \begin{array}{lll}
        u&=0, & \forall x=1, \\
        u&=1, & \forall x=0, \\
        \dfrac{\partial u}{\partial n}&=0, & \forall y=0 \text { and } y=1.
    \end{array}
\end{equation}
Furthermore, to capture the associated uncertainty, the diffusion coefficient $a(x,y)$ is modeled as a log-normal random field, which can be expressed as follows:

\begin{equation}
    \log a(x,y) \sim \operatorname{GP}\left(a(x,y) \mid m(x,y), \mathcal{K}\left((x, y),\left(x^{\prime}, y^{\prime}\right)\right)\right),
\end{equation}
where the mean function $m(x,y) = 0$, and the covariance function $\mathcal{K}\left((x, y),\left(x^{\prime}, y^{\prime}\right)\right)$ is modeled as a GRF. The covariance kernel can be represented as follows:

\begin{equation}
    \mathcal{K}\left((x, y),\left(x^{\prime}, y^{\prime}\right)\right) = \exp \left( \frac{-\left(x-x^{\prime}\right)^{2}}{2 l_{1}^{2}}+\frac{-\left(y-y^{\prime}\right)^{2}}{2 l_{2}^{2}} \right).
\end{equation}
The covariance kernel's parameter $l_{1}$ and $l_{2}$ are set equal to $0.25$. Our aim with this problem is to learn the operator mapping from the logarithm of the diffusion coefficient to the solution of the PDE, which could be mathematically written as follows:

\begin{equation}
    \mathcal{D}: \log(a(x, y)) \mapsto u(x, y).
\end{equation}
 However, in the present case, this map is learned using multi-fidelity data. Furthermore, to generate the LF and HF data, a finite volume (FV) solver provided by Tripathy and Bilionis \cite{tripathy2018deep} is modified, and the stochastic steady state equation is solved on grids of sizes $6\times6$ and $32\times32$ for LF and HF cases, respectively.

\subsubsection{Surrogate construction using MF-WNO}\label{S:341}
\begin{figure}[ht!]
\centering
\subfigure[]{\label{subfig:lab91}\includegraphics[width=0.49\textwidth]{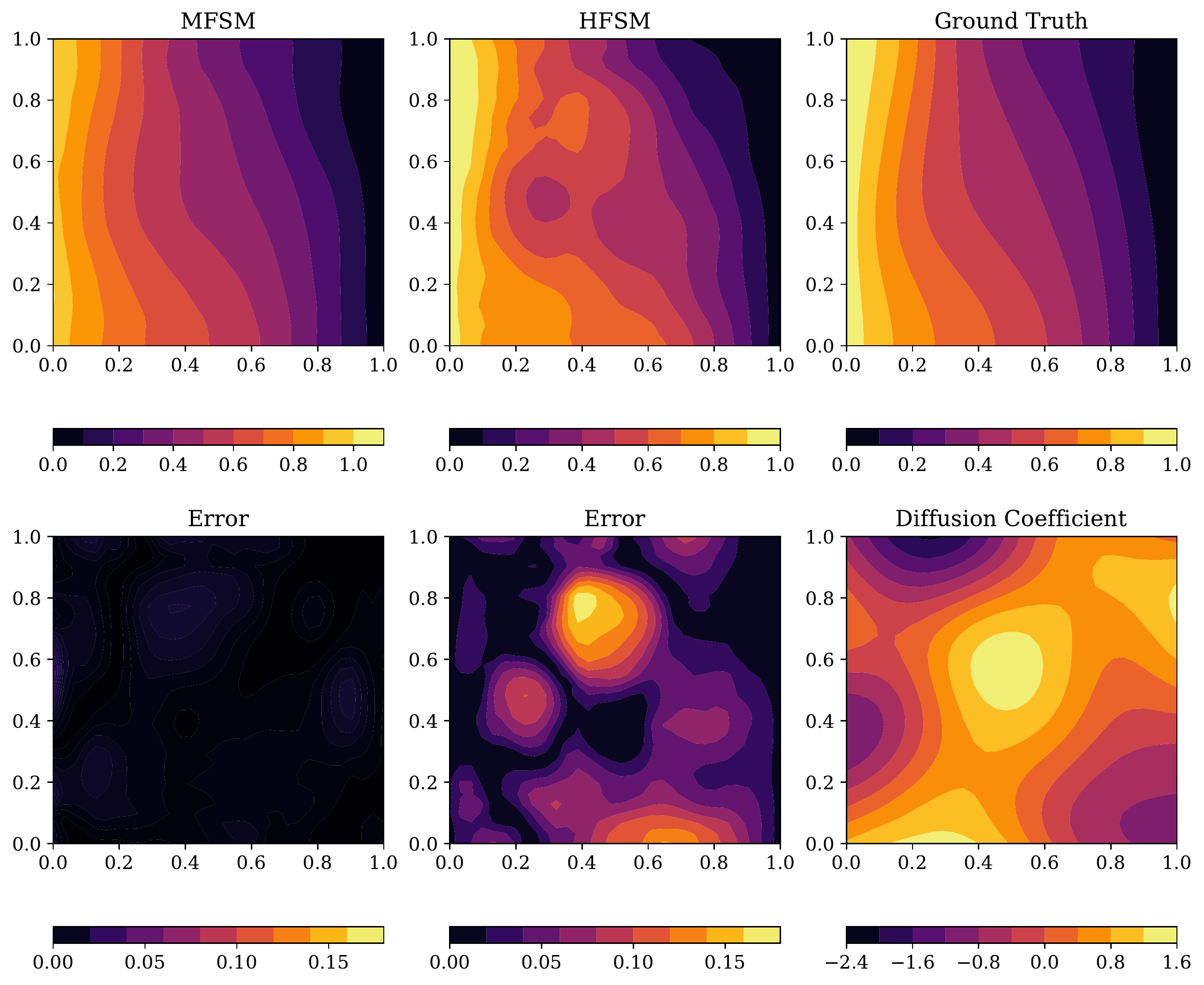}}
\subfigure[]{\label{subfig:lab92}\includegraphics[width=0.49\textwidth]{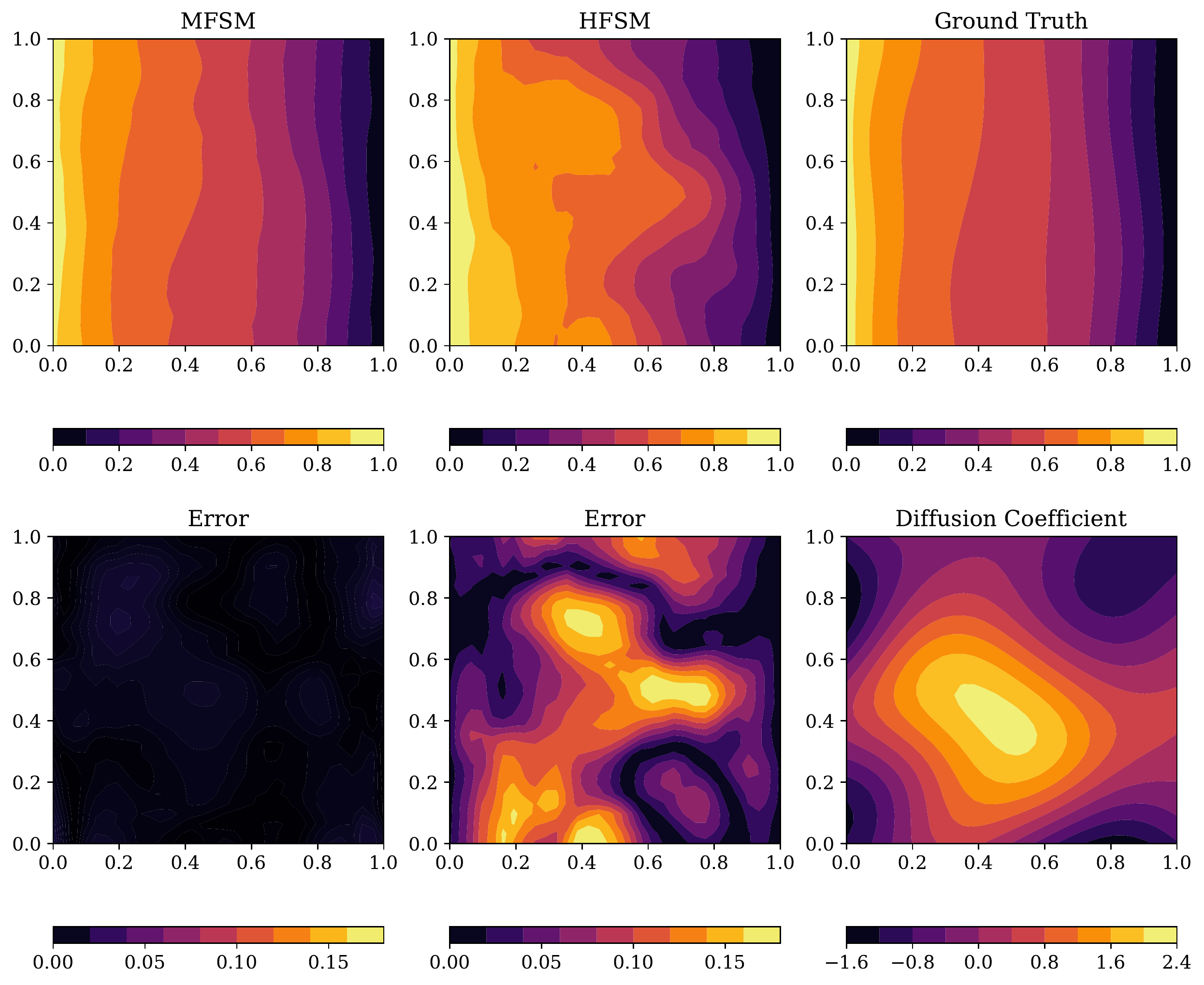}}
\subfigure[]{\label{subfig:lab93}\includegraphics[width=0.49\textwidth]{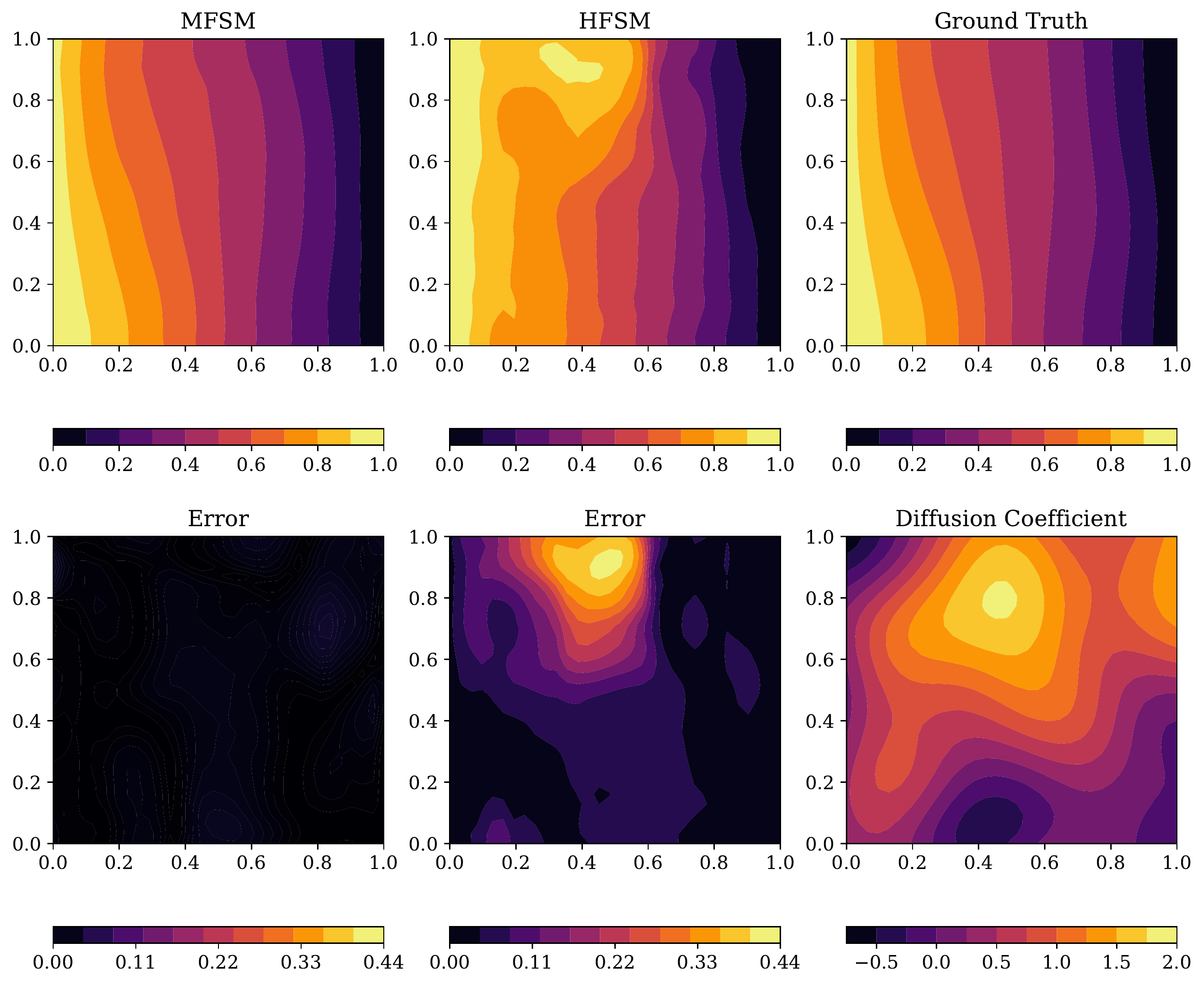}}
\subfigure[]{\label{subfig:lab94}\includegraphics[width=0.49\textwidth]{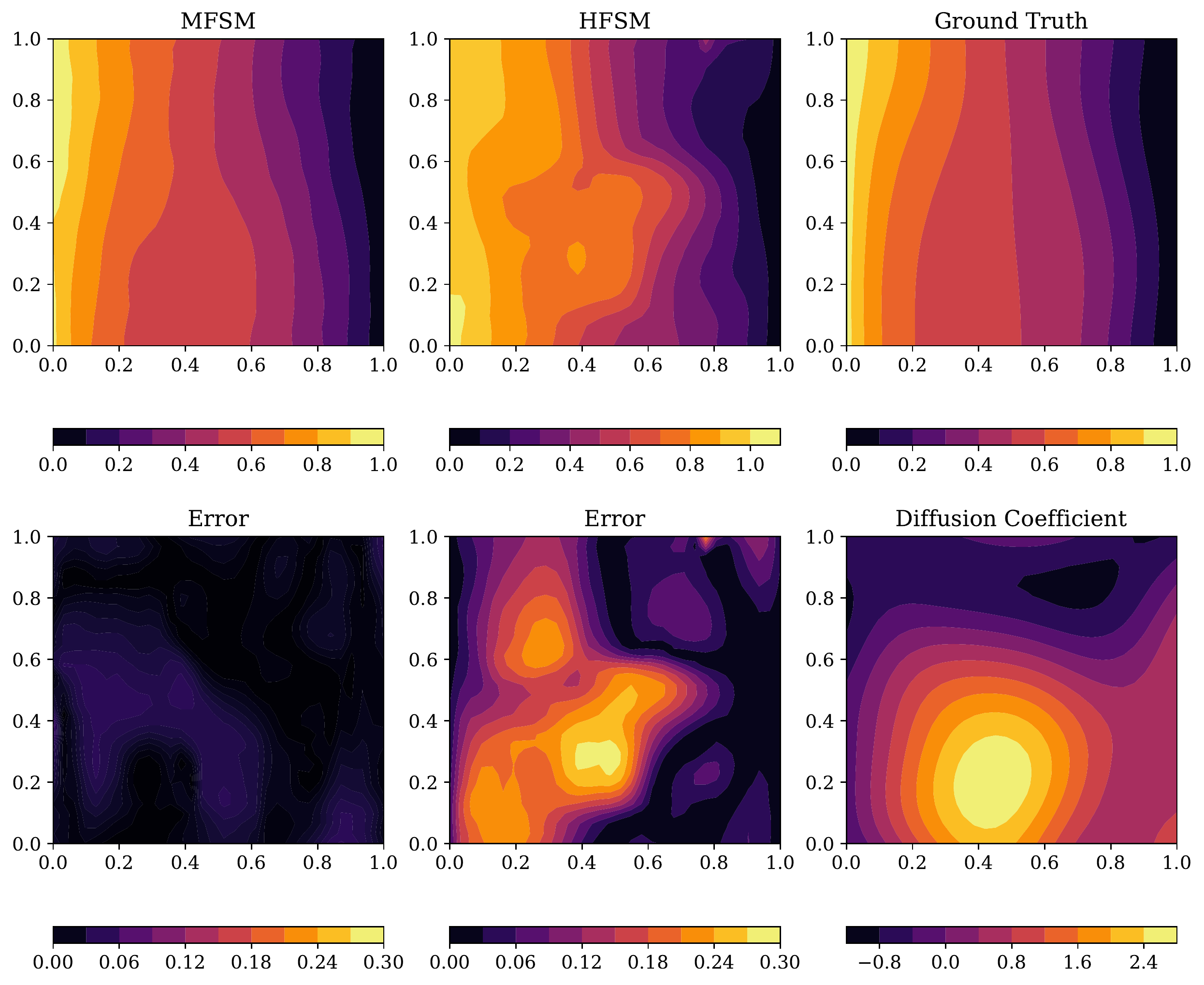}}
\caption{Comparison of exact HF solution with the solution predicted by MFSM-WNO and predictions from HFSM-WNO along with respective absolute errors and pictorial representation of logarithm of diffusion coefficient for training dataset sizes of \textbf{(a)} $25$, \textbf{(b)} $15$, \textbf{(c)} $8$, and \textbf{(d)} $5$ for the stochastic heat equation.}
\label{fig:9}
\end{figure}

We train surrogate models (or MFSMs and HFSMs, respectively) using dataset sizes of $25$, $15$, $8$, and $5$ using bi-fidelity data with MF-WNO and MF-DeepONet and only HF data with vanilla WNO. In order to visually compare the outputs from MFSM-WNO and HFSM-WNO for different training dataset sizes, a pictorial presentation of the predicted solutions, errors, actual HF solutions or ground truth, and associated input field, i.e., the logarithm of diffusion coefficient, is provided in Fig. \ref{fig:9}. Furthermore, for all the different dataset sizes, the MSE values for the prediction of MFSMs and HFSM on unseen test datasets are made available in Table \ref{tab:5}.\par
\begin{table}[htbp!]
    \centering
    \caption{MSE error between exact HF solution and predictions from MFSM-WNO, HFSM-WNO, and MFSM-DeepONet on unseen test dataset for different training dataset sizes for the stochastic heat equation.}
    \label{tab:5}
    \begin{tabular}{m{3cm}m{2cm}m{2cm} c} 
    \toprule
    \multirow{2}{*}{$\mathcal{Q}_{train}$ Size} &\multicolumn{3}{c}{MSE}\\ \cline{2-4}
    & MFSM-WNO & HFSM-WNO & MFSM-DeepONet \\ [0.5ex] 
    \midrule
    \vspace{0.2em}
    $25$ & 1.4001 $\times 10^{-4}$ & 3.0188 $\times 10^{-3}$ &2.7084 $\times 10^{-4}$\\ 
    $15$ & 2.7293 $\times 10^{-4}$ & 5.5890 $\times 10^{-3}$ &4.6942 $\times 10^{-4}$\\ 
    $8$ & 4.5502 $\times 10^{-4}$ & 1.3692 $\times 10^{-2}$  &1.8211 $\times 10^{-3}$\\ 
    $5$ & 7.2992 $\times 10^{-4}$ & 2.5784 $\times 10^{-2}$  &4.3650 $\times 10^{-3}$\\ 
    \hline
    $MSE_{LF}$ & \multicolumn{3}{c}{1.3793 $\times 10^{-3}$}\\
    \bottomrule
    \end{tabular}
    
\end{table}
After looking at Table \ref{tab:5}, it becomes obvious that MFSM-WNO outperforms HFSM-WNO and MFSM-DeepONet for all training dataset sizes. However, MFSM-DeepONet's predictions are very close to MFSM-WNO in terms of accuracy. Furthermore, it can be noticed that there is two order of magnitude decrease in MSE for MFSM-WNO as compared to HFSM-WNO for training dataset sizes $8$ and $5$, while there is a decrease in error by an order of magnitude for sizes $25$ and $15$. Moreover, Fig. \ref{fig:9} clearly shows the doubtless superiority of MFSM-WNO when compared to HFSM-WNO for all sizes of datasets used for model training purposes. Finally, the time needed by MFSM-WNO for predicting solutions for $200$ samples for the current example is $0.40$ seconds.

\subsubsection{Comparison of LF-WNO and LF solver for the supply of low-fidelity data}
\begin{figure}[htbp!]
    \centering
    \subfigure[]
    {\label{subfig:lab121}\includegraphics[width=0.4\textwidth]{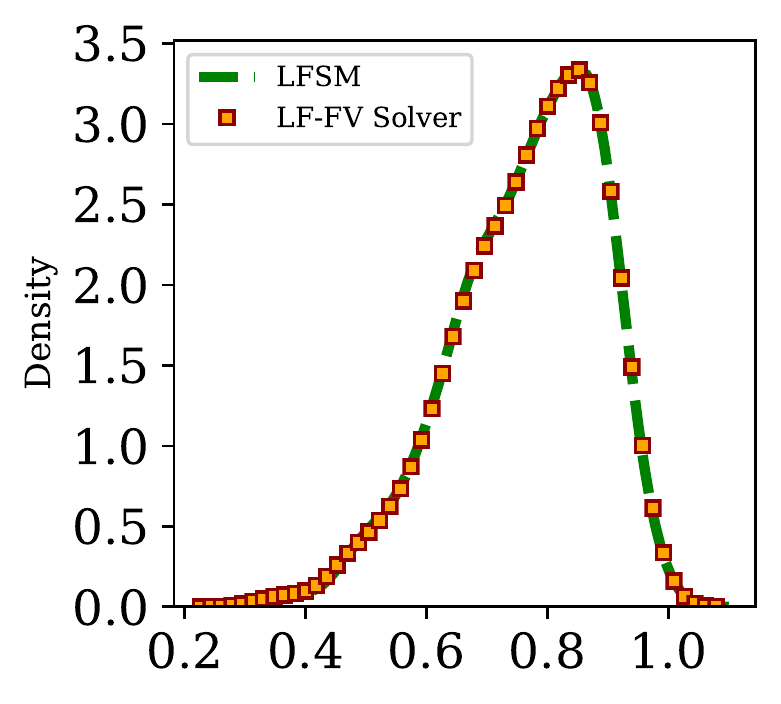}}
    \subfigure[]
    {\label{subfig:lab122}\includegraphics[width=0.4\textwidth]{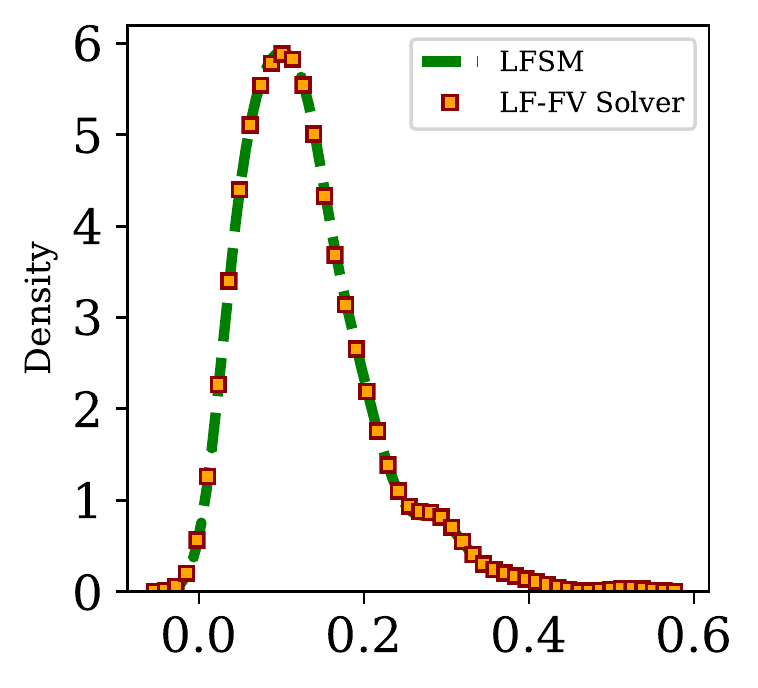}}
    \caption{Comparison of PDE solution density at spatial grid points \textbf{(a)} $(x,y) =  (0.203, 0.515)$ and \textbf{(b)} $(x,y) = (0.859, 0.391)$ between LFSM predictions and the actual solution from LF-FV solver.}
    \label{fig:12}
\end{figure}

It is possible that although LF data is present, we might not have access to the LF solver. In such cases, it would be challenging to generate new LF solutions that might be required for predictions of HF solutions using HF-WNO for unavailable input functions. However, in these cases, we could always turn towards training a surrogate using LF-WNO and the available data. Therefore, to establish that LF-WNO can easily replace an LF solver, in this section, we compare the utilization of a low fidelity surrogate model (LFSM), trained using LF-WNO on a training dataset of size $3000$, with the LF-solver as the source for low-fidelity solution data, which would then be used for residual learning and input supplementation in HF-WNO. To evaluate the accuracy of LFSM, firstly, the MSE between LFSM predictions and actual LF-solution for a given test set. The said MSE is found to be $3.2044\times 10^{-5}$. Furthermore, we evaluate the MSE between the LF-solution predicted from LFSM for an unseen test set and the corresponding exact HF solution. We draw a comparison by performing a similar evaluation for corresponding LF-solutions from LF-solver. The MSE obtained for LFSM predictions and solutions from LF-solver with respect to exact HF-solution are $1.3270\times 10^{-3}$ and $1.3081\times 10^{-3}$, respectively. Clearly, even for a training dataset size of $3000$, the MSE values on the test set indicate that the surrogate predicted and actual LF solutions very closely match each other. The success of LFSM in accurately mimicking the LF solution is further illustrated by the pdf plots at points $(x,y) =  (0.203, 0.515)$ and $(x,y) = (0.859, 0.391)$ for LFSM and actual solution from the LF-solver in Fig. \ref{fig:12}.    
Furthermore, we also develop an MFSM on a training dataset of size $25$ similar to section \ref{S:341}. However, unlike the MFSM in section \ref{S:341}, where the LF solver was directly used for the supply of LF solution to HF-WNO, here we use the LF solution predictions from LF-WNO as inputs to HF-WNO. The resulting MSE obtained for MFSM on a test set is $1.5392\times 10^{-4}$ compared to the MSE value of $1.4001\times 10^{-4}$ for MFSM in section \ref{S:341}. This again indicates that due to the ability of LFSM to become highly accurate with sufficient training data, the result obtained from MFSM trained using LF data from LFSM is similar to that from MFSM trained using LF data directly from the LF solver. It should be noted that the accuracy of LFSM would keep on increasing with a further increase in training dataset size.  
\begin{figure}[htbp!]
    \centering
    \subfigure[]
    {\label{subfig:lab131}\includegraphics[width=0.4\textwidth]{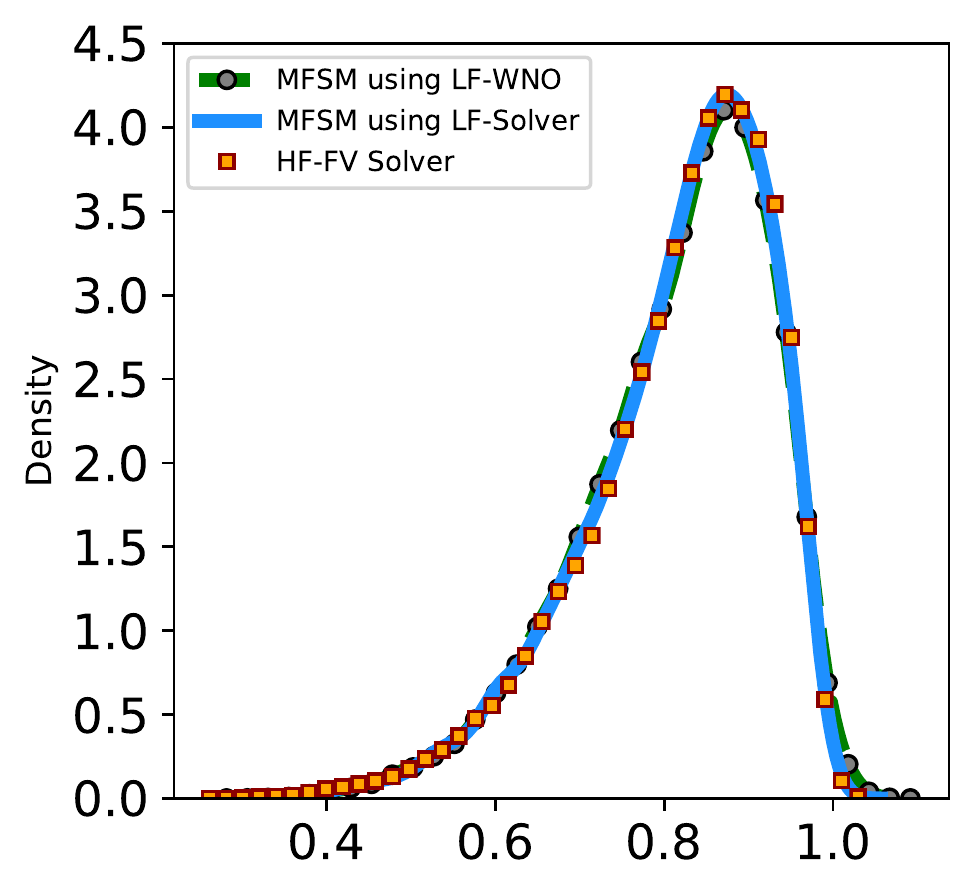}}
    \subfigure[]
    {\label{subfig:lab132}\includegraphics[width=0.4\textwidth]{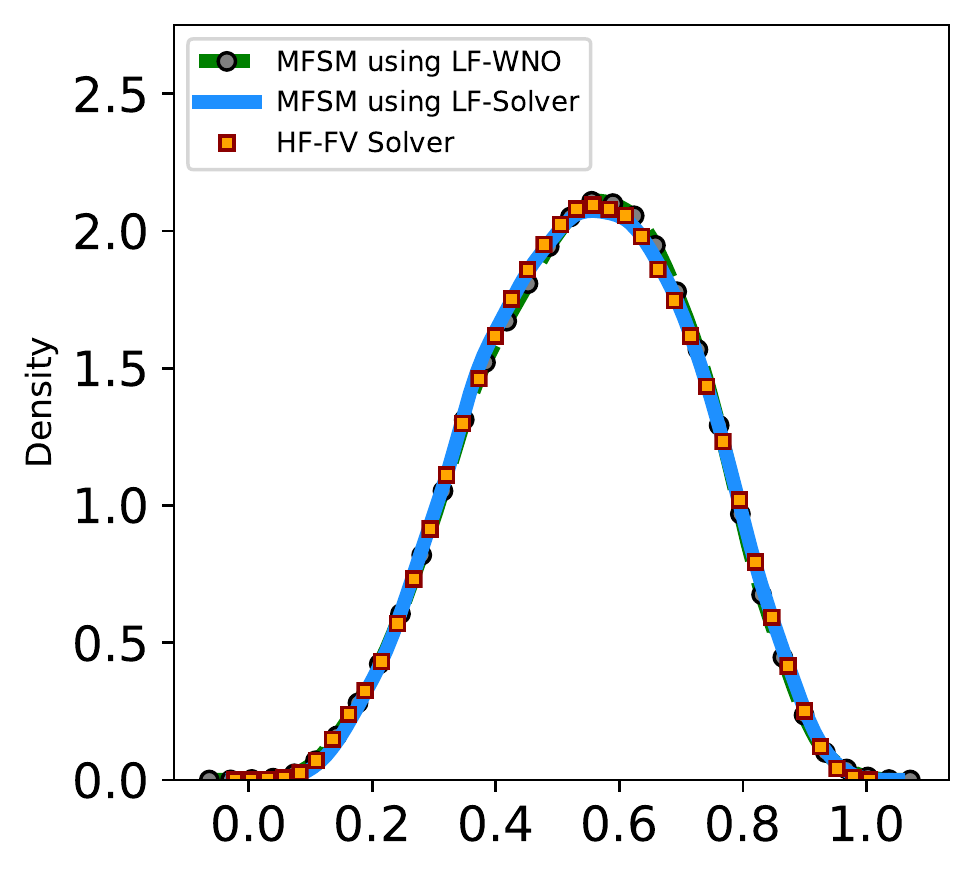}}
    \caption{Comparison of PDE solution density at spatial grid points \textbf{(a)} $(x,y) =  (0.172, 0.141)$ and \textbf{(b)} $(x,y) = (0.453, 0.484)$ among the LF-WNO based MFSM's predictions, LF solver based MFSM's predictions, and the actual solution from HF-FV solver.}
    \label{fig:13}
\end{figure}
In view of all the results presented in this section, it can be safely stated that an LF-WNO can accurately and efficiently replace LF solver.

\subsection{Problem Set IV: Unsteady multi-fidelity learning }
\subsubsection{2-dimensional Allen Cahn equation}
The 2-dimensional Allen-Cahn equation is a type of reaction-diffusion equation that finds pervasive utility in the field of chemical reactions and the modeling of phase separation phenomena in multicomponent alloys. Specifically, this PDE is employed to elucidate the intricate dynamics governing the evolution of interfaces and the behavior of materials undergoing phase separation in two-dimensional spatial domains. Its mathematical formulation can be given as
\begin{equation}
\partial_t u(x, y, t)=\epsilon \Delta u(x, y, t)+u(x, y, t)-u(x, y, t)^3, \;x, y \in(0,3), \;t \in[0,10],
\end{equation}
where $\epsilon \in \mathbb{R}^{+*}$ is a positive real constant that controls the extent of diffusion. Further, the boundaries of the 2-dimensional domain in the current problem are periodic, and the value of $\epsilon = 1 \times 10^{-3}$. Finally, the initial condition $u(x, y, 0)=u_0(x, y) \;x, \;y \in(0,3)$ is sampled from a random field using the following kernel
\begin{equation}
   \mathcal{K}(x, y)=\tau^{(\alpha-1)}\left(\pi^2\left(x^2+y^2\right)+\tau^2\right)^{\frac{\alpha}{2}}, 
\end{equation}
where the kernel parameters are set as $\tau=15$ and $\alpha=1$. We utilize a grid size of $65 \times 65$ and generate the ground truth data comprising of snapshots of the evolution of $80$ different initial conditions till the time $t=10$ s. We use spectral methods for obtaining the discrete approximations of the differential operators on the right-hand side of the equation. However, we march forward in time in the physical space (not the spectral space) using the forward Euler temporal integrator with a time-step size $\Delta t = 0.04$ s. Finally, we obtain the training and testing data by subsampling the evolution trajectories at temporal locations separated by $5\Delta t$, i.e., we increase the time step 5 times for our neural operator surrogate. Also, we systemically reduce the number of spatiotemporal evolution (SE) trajectories\footnote{Spatiotemporal evolution trajectories refers to the complete discrete spatiotemporal dynamics for different initial conditions till $10$ s obtained from the numerical solver for the PDE under consideration.} used for training our surrogates from $80$ to $40$. Furthermore, the test data comprises data equivalent to $10$ evolution trajectories and is sampled from the combined training and testing dataset.
In addition, in this case, the goal of HFSM is to learn the operator mapping from  $u_t(x, y)$ to $u_{t+5\Delta t}(x, y, t)$. Thus, the learning framework in the training phase becomes a next-step teacher-forcing type. A related work can be found in \cite{liu2022predicting}. \par
\begin{figure}[htbp!]
    \centering
    \includegraphics[width=0.9\textwidth]{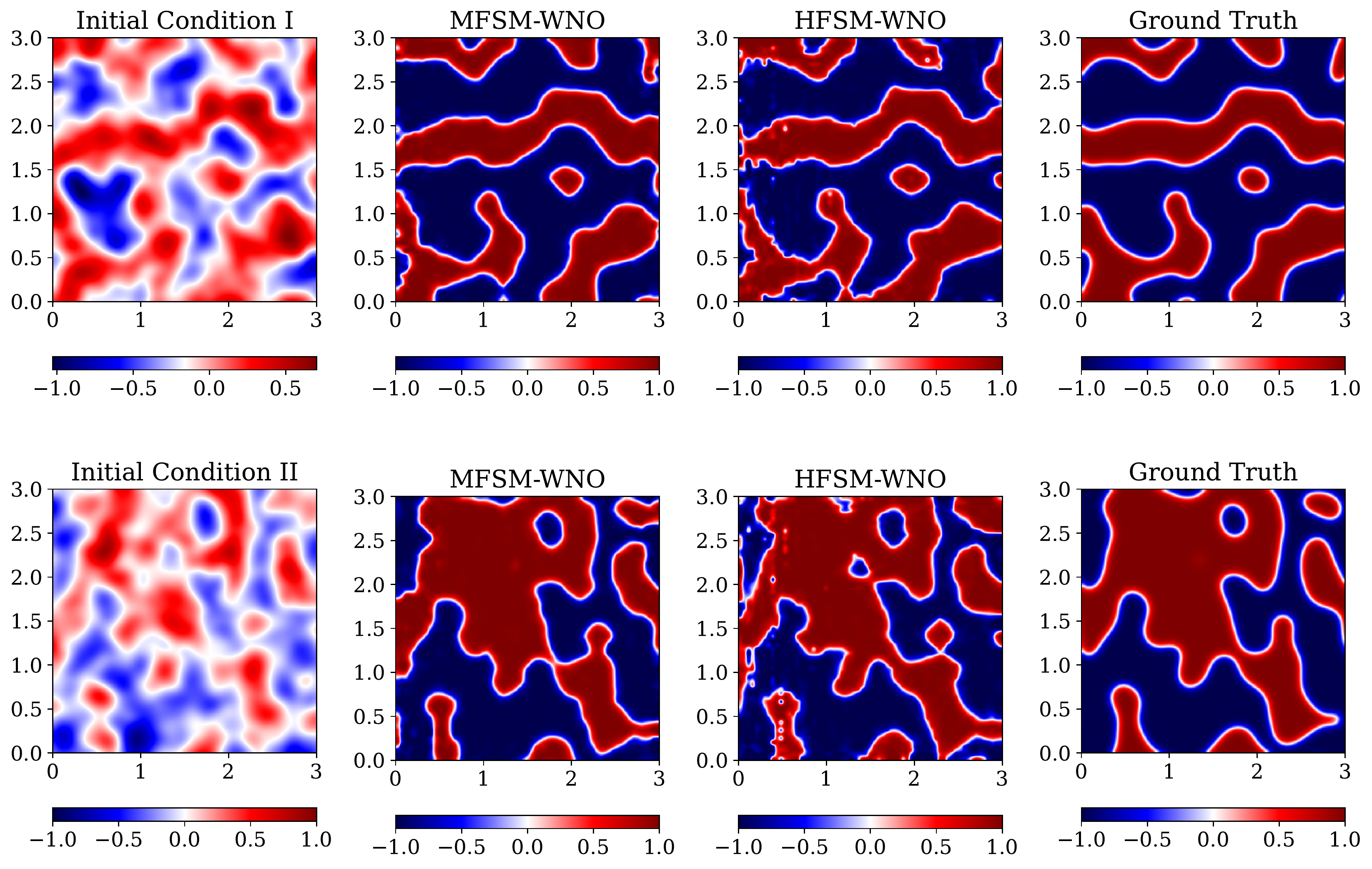}
    \caption{Comparison of predicted fields from MFSM-WNO and HFSM-WNO at t = $10$ s, which were obtained using a temporal rollout from two different initial conditions, with ground truth. Each row contains results for IC in the corresponding first column.}
    \label{fig:14}
\end{figure}

Finally, the multi-fidelity learning strategy is kept the same. For MFSM-WNO, additionally, we augment the input $u_t(x, y)$ with $\bar u_{t+5\Delta t}(x, y, t)$, where $\bar u_{t+5\Delta t}$ is obtained by solving $\bar u_t(x, y)$ (twice spatially subsampled $u_t(x, y)$ or velocity at the current time step) using the numerical solver and a time step 5 times larger than the ground truth simulation. Also, similarly, we subtract $\bar u_{t+5\Delta t}(x, y, t)$ from $u_{t}(x, y, t)$ to formulate our residual operator learning problem. After training, it was found that the predictive MSE for MFSM-WNO on the test dataset is around one order of magnitude better than HFSM-WNO, and the MSE values are tabulated in Table \ref{tab:8}. Also, the performance of MFSM-WNO is also assessed in predicting entire spatiotemporal dynamics for $100$ unseen initial conditions (IC) till $10$ s. A comparative assessment of MSE values for MFSM-WNO and HFSM-WNO for this task is present in Table \ref{tab:9}. Finally, a pictorial comparison of evolved spatiotemporal dynamics at time t = $10$ s for some randomly selected samples from the unseen $100$ ICs for the models in contention is provided in Figure \ref{fig:14}. In general, it is found that MFSM-WNO outperforms HFSM-WNO in all the conducted experiments.
\begin{table}[htbp!]
    \centering
     \caption{MSE error between exact HF  and predictions from MFSM-WNO and HFSM-WNO for complete trajectory prediction test dataset for different number of total trajectories present in the combined training and testing dataset for unsteady Allen-Cahn equation.}
    \label{tab:8}
    \begin{tabular}{c m{2cm}m{2cm}} 
    \toprule
    \multirow{2}{*}{Total SE trajectories in dataset} &\multicolumn{2}{c}{MSE}\\ \cline{2-3}
    & MFSM-WNO & HFSM-WNO \\ [0.5ex] 
    \midrule
    \vspace{0.2em}
    $80$ & 2.9312 $\times 10^{-5}$ & 1.3158 $\times 10^{-4}$ \\ 
    $40$ & 4.2087 $\times 10^{-5}$ & 3.1563 $\times 10^{-4}$  \\ 
    \hline
    \vspace{0.2em}
    $\text{Average MSE}_{LF}$ & \multicolumn{2}{c}{3.055 $\times 10^{-3}$}\\
    \bottomrule
    \end{tabular}
    
\end{table}

\begin{table}[htbp!]
    \centering
    \caption{MSE error between exact HF SE trajectories and predicted SE trajectories from MFSM-WNO and HFSM-WNO trained using different dataset sizes for 100 unseen ICs with the unsteady Allen-Cahn equation as the PDE in consideration.}
    \label{tab:9}
    \begin{tabular}{c m{2cm}m{2cm}} 
    \toprule
    \multirow{2}{*}{Total trajectories in dataset} &\multicolumn{2}{c}{MSE}\\ \cline{2-3}
    & MFSM-WNO & HFSM-WNO \\ [0.2ex] 
    \midrule
    \vspace{0.2em}
    $80$ & 0.0271  & 0.0779 \\ 
    $40$ & 0.0486  & 0.1123  \\ 
    \bottomrule
    \end{tabular}
\end{table}

\section{Conclusion}\label{S:4}
In this work, we proposed a novel framework for WNO that allows accurate and efficient learning from a multi-fidelity dataset. The novel framework was developed by coupling separate WNO networks together with the help of residual operator learning and supplementation of input. The proposed approach was based upon the utilization of an inexpensive to generate LF dataset of large size together with an expensive to generate HF dataset of small size.

The performance of the proposed framework was assessed on several problems, including artificial benchmarks and complex PDEs, and the results consistently showed that the surrogate model constructed using MF-WNO provides predictions that are at least one order of magnitude and, in many cases, multiple orders of magnitude more accurate than both the LF solution and predictions made by vanilla WNO with only HF dataset. Also, MF-WNO performed better as compared to MF-DeepONet for all cases. In addition, the propounded approach does well in uncovering the correlations between the LF and HF data, even when they are nonlinear and complex. Furthermore, the framework successfully tackles problems with an irregular domain. Also, from the results obtained from the stochastic heat equation problem, it is observed that our framework is robust and accurate for surrogate modeling for stochastic PDEs in the low data limit. Also, a bi-fidelity framework was developed for surrogate modeling of a nonlinear and an unsteady PDE, and it faired well in its assessment with respect to HF-only surrogate for the unsteady PDE. However, a pertinent direction for future research could be the extension of the current framework to a probabilistic regime.

\section{Acknowledgement}
T. Tripura acknowledges the financial support received from the Ministry of Human Resource Development (MHRD), India in the form of the Prime Minister's Research Fellows (PMRF) scholarship. S. Chakraborty acknowledges the financial support received in part from Science and Engineering Research Board (SERB) through grant no. SRG/2021/000467 and in part as a seed grant from IIT Delhi.


\end{document}